\documentclass[10pt,twocolumn,letterpaper]{article}

\usepackage{icb}
\usepackage{times}
\usepackage{epsfig}
\usepackage{graphicx}
\usepackage{amsmath}
\usepackage{amssymb}
\usepackage{multirow}
\usepackage{subcaption}

\usepackage{epstopdf}
\usepackage[breaklinks=true,colorlinks,bookmarks=false]{hyperref}

\usepackage[utf8]{inputenc}
\usepackage[T1]{fontenc}

\usepackage{graphicx} 
\graphicspath{{./figures/}}
\DeclareGraphicsExtensions{.pdf,.png,.jpg}

\usepackage{color, colortbl}
\definecolor{claA}{rgb}{1.00,1.00,1.00}
\definecolor{claB}{rgb}{0.90,0.90,0.90}
\definecolor{claC}{rgb}{0.80,0.80,0.80}
\definecolor{LightCyan}{rgb}{0.88,1,1}

\usepackage[printwatermark]{xwatermark}
\newwatermark*[allpages,color=red,angle=0,scale=1,xpos=0,ypos=370pt,fontsize=7pt]{
The content of this paper was published in ICB, 2019. This ArXiv version is from before the peer review. Please, cite the following paper:
\\Ž. Emersic, A. Kumar S. V., B. S. Harish, W. Gutfeter, J. N. Khiarak, A. Pacut, E. Hansley, M. Pamplona Segundo, S. Sarkar, H. Park, \\
G. Pyo Nam, I. J. Kim, S. G. Sangodkar, U. Kacar, M. Kirci, L. Yuan, J. Yuan, H. Zhao, F. Lu, J. Mao,X. Zhang, D. Yamah, \\
F. I. Eyiokur, K. B. Ozler, H. K. Ekenel, D. Paul Chowdhury, S. Bakshi, P. K. Sa, B. Majhi, P. Peer, V. Struc: \\“The Unconstrained Ear Recognition Challenge 2019”, International Conference on Biometrics, IEEE, 2019
}

\icbfinalcopy 

\ificbfinal\fi
\begin{document}


\title{The Unconstrained Ear Recognition Challenge 2019\\ -- ArXiv Version With Appendix}

\author{
\begin{minipage}{1\textwidth}
\small
\renewcommand{\baselinestretch}{1.15}
\centering
Ž.~Emeršič$^1$, A.~Kumar S.~V.$^2$,  B.~S.~Harish$^3$, W.~Gutfeter$^4$,  J.~N.~Khiarak$^5$,  A.~Pacut$^5$,  E.~Hansley$^6$,  M.~Pamplona Segundo$^7$, S.~Sarkar$^6$,   
\small H.~Park$^8$, G.~Pyo Nam$^8$,  I.~J.~Kim$^8$,  S.~G.~Sangodkar$^9$,  U.~Kacar$^{10}$,  M.~Kirci$^{10}$,  L.~Yuan$^{11}$, J.~Yuan$^{11}$, H.~Zhao$^{11}$, F.~Lu$^{11}$, J.~Mao$^{11}$, X.~Zhang$^{11}$,  D.~Yaman$^{10}$,  F.~I.~Eyiokur$^{10}$, K.~B.~\"{O}zler$^{10}$, H.~K.~Ekenel$^{10}$, D.~Paul~Chowdhury$^{12}$, S.~Bakshi$^{12}$, P.~K.~Sa$^{12}$, B.~Majhi$^{12}$, P.~Peer$^1$, V.~Štruc$^1$\\
\small 
$^1$University of Ljubljana (UL, SI), $^2$Nitte Mahalinga Adyanthaya Memorial Institute of Technology (NMAM IT, IN), $^3$Jagadguru Sri Shivarathreeshwara Science \& Technology University (JSS STU, IN), $^4$Research an Academic Computer Network (NASK, PL), $^5$Warsaw University of Technology (WUT, PL), $^6$University of South Florida (USF, USA), $^7$Federal University of Bahia (UFBA, BR), $^8$Korea Institute of Science and Technology (KIST, KR),  $^9$Indian Institute of Technology Bombay (IITB, IN), $^{10}$Istanbul Technical University (ITU, TR), $^{11}$University  of  Science  and  Technology  Beijing (USTB, CN), $^{12}$National Institute of Technology Rourkela (NITR, IN)\\ 
\end{minipage}}

\maketitle

\begin{abstract}

This paper presents a summary of the 2019 Unconstrained Ear Recognition Challenge (UERC), the second in a series of group benchmarking efforts centered around the problem of person recognition from ear images captured in uncontrolled settings.
The goal of the challenge is to assess the performance of existing ear recognition techniques on a challenging large-scale ear dataset and to analyze performance of the technology from various viewpoints, such as generalization abilities to unseen data characteristics, sensitivity to rotations, occlusions and image resolution and performance bias on sub-groups of subjects, selected based on demographic criteria, i.e. gender and ethnicity. Research groups from $12$ institutions entered the competition and submitted a total of $13$ recognition approaches ranging from descriptor-based methods to deep-learning models. The majority of submissions focused on ensemble based methods combining either representations from multiple deep models or hand-crafted with learned image descriptors. Our analysis shows that methods incorporating deep learning models clearly outperform techniques relying solely on hand-crafted descriptors, even though both groups of techniques exhibit similar behaviour when it comes to robustness to various covariates, such presence of occlusions, changes in (head) pose, or variability in image resolution. The results of the challenge also show that there has been considerable progress since the first UERC in 2017, but that there is still ample room for further research in this area. 
\end{abstract}

\section{Introduction}


%
Biometric ear recognition refers to the task of recognizing people from ear images using computer vision and machine learning techniques. The interest in this field is driven by the appealing characteristics of the human ear when used in automated recognition systems, such as the ability to capture images from a distance and without explicit cooperation of the subjects one is trying to recognize, the ability to distinguish identical twins~\cite{Sim2012ICPR} and the potential to supplement other biometric modalities (e.g., faces) in multi-modal biometric systems~\cite{theoharis2008unified, xu2007feature}.

\begin{figure}
\centering
\begin{subfigure}{.49\columnwidth}
  \centering
  \includegraphics[width=0.99\columnwidth]{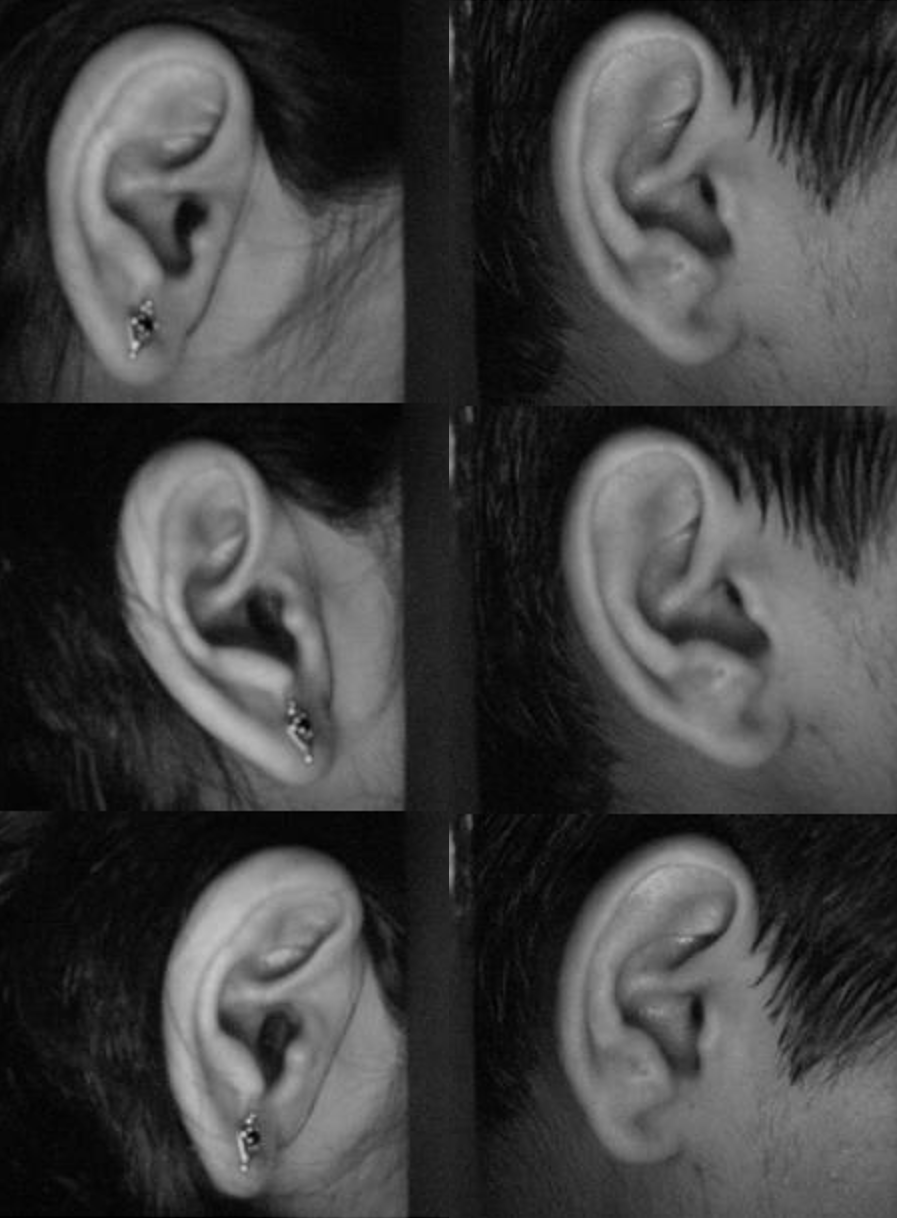}
  \label{fig:sub1}\vspace{-3mm}
\end{subfigure}%
\hfill
\begin{subfigure}{.49\columnwidth}
  \centering
  \includegraphics[width=0.99\columnwidth]{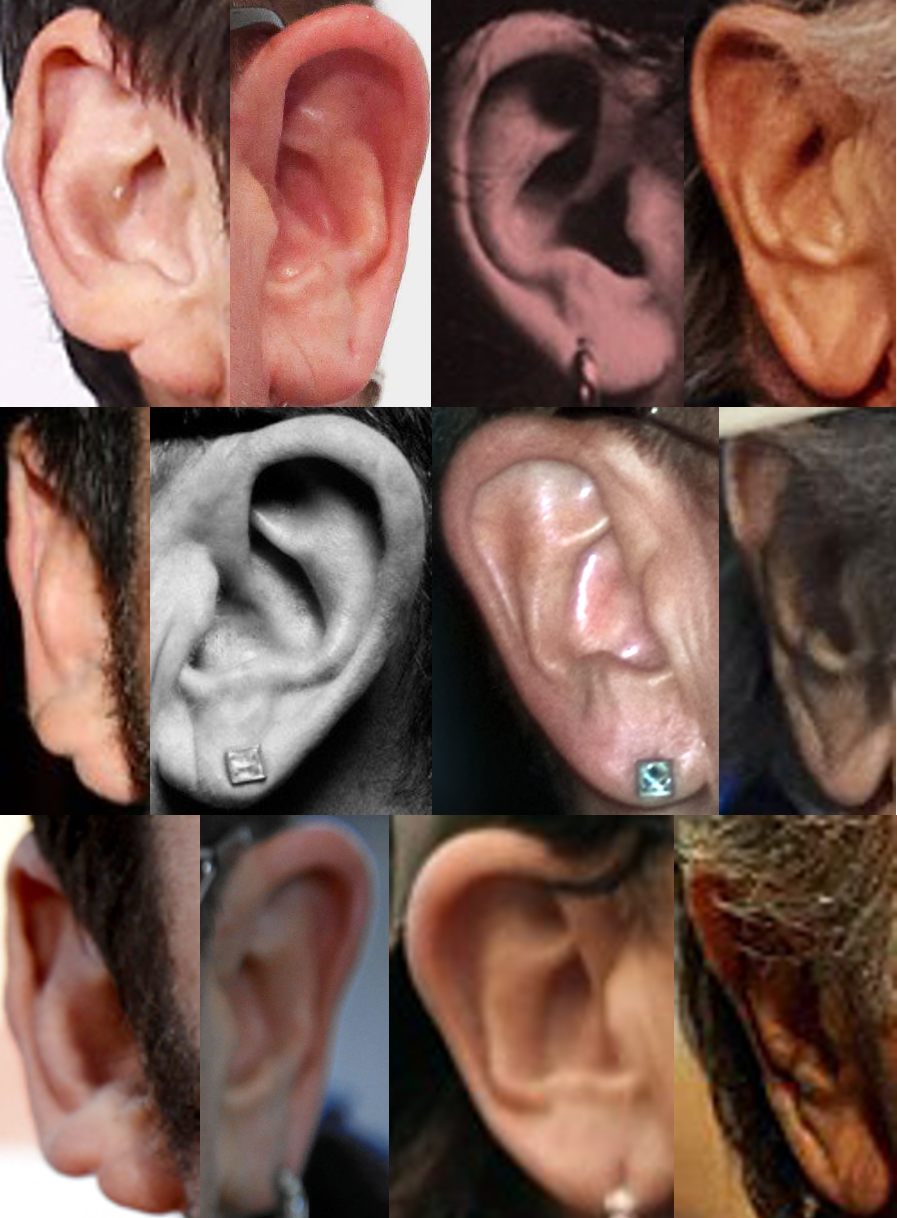}
  \label{fig:sub2}\vspace{-3mm}
\end{subfigure}
\caption{Early research on ear recognition focused mostly on constrained laboratory-like settings - see sample images on the left~\cite{Kumar2012}. Recent work is looking at more realistic images captured in unconstrained conditions - see examples from the database used for UERC~\cite{ncaa} on the right.  UERC 2019 aims to benchmark (in a group effort) existing ear recognition techniques  
on such unconstrained data. Each column shows images of one subject. 
\vspace{-2mm}}
\label{datasetComparison}
\end{figure}

%

Research on ear recognition has long been focused on constrained laboratory-like settings, where the  variability of the captured ear images was limited and not really representative of real-world settings. The main sources of appearance variability with the ear datasets used during this period were minute differences in head rotation, presence (or absence) of ear accessories and minor changes in illumination conditions,  
as illustrated by the sample images on the left side of Figure~\ref{datasetComparison}. While these datasets contributed to early developments in the field, they did not account for challenging real-life conditions where the appearance of ears is significantly affected by blur, illumination, occlusion, and view-direction changes as well other nuisance factors, as shown on the right side of Figure \ref{datasetComparison}. New datasets were needed that would better reflect the image variability encountered in unconstrained (real-world) settings. To address this gap, the Annotated Web Ears (AWE) dataset was introduced in~\cite{Survey2017}, followed by related unconstrained ear datasets shortly after~\cite{pedced,ncaa, emervsivc2017unconstrained,zhang2017ear}. Similarly to recent image collections from other areas of computer vision~\cite{cao2018vggface2,huang2008labeled,parkhi2015deep}, these ear datasets were not captured in laboratory environments, but were gathered from the internet using automatic (and/or semi-automatic) acquisition techniques. Because of the web-based collection approach, ear images from these datasets better reflect the  range of possible appearance variability encountered in real-world settings, but consequently also pose a considerably greater challenge to existing recognition techniques. 

To analyze how existing ear recognition techniques perform with such difficult images, the first Unconstrained Ear Recognition Challenge (UERC) was organized in 2017 as part of the International Joint Conference on Biometrics (IJCB). The challenge was conducted on a large-scale dataset of ear images 
and showed that the recognition performance in unconstrained settings is far from the near-perfect recognition rates commonly reported  on constrained laboratory datasets. In the most difficult (identification) experiment involving a gallery set of several thousand subjects, the best performing algorithm achieved a relatively modest rank-$1$ recognition rate of little above $20\%$, leaving ample room for further research and improvements. Moreover, while certain aspects of the submitted algorithms were studied at UERC 2017 (i.e., impact of head rotations, scalability, etc.), important research questions that could offer insight into potential issues with the existing recognition techniques and point to future research directions were left unanswered, e.g.: How do ear recognition techniques perform across different image resolutions in unconstrained settings? How sensitive are existing techniques to the presence of occlusions and ear accessories? 
Do existing recognition approaches exhibit a performance bias when presented with images of either male or female subjects? How do recognition techniques generalize to ear image data with different characteristics? 

To answer these and related questions, the second Unconstrained Ear Recognition Challenge (UERC) was organized in the scope of the 2019 IAPR International Conference on Biometrics (ICB). UERC 2019 is a follow up on the first ear recognition challenge and with $12$ participating institutions the biggest group effort so far aimed at evaluating the state of technology in the field of unconstrained ear recognition. The participating research groups submitted a total of $13$ algorithms that were tested on a common dataset of ear images using a predefined experimental protocol. To allow for comparisons with the results from UERC 2017, the same dataset and protocol was used for the main part of UERC 2019, but a new, sequestered datasets was added to evaluate the generalization abilities of the submitted algorithms. The combined research effort of all participating groups resulted in the following contributions: 


\begin{itemize}
    \item A comparative performance evaluation of ear recognition technology involving $13$ recognition techniques and comprehensive experiments on an unconstrained large-scale dataset of ear images.
    \item An in-depth analysis of the sensitivity of ear recognition techniques to different covariates known to affect performance, e.g., image resolution, head rotation, and presence of occlusions. 
    \item An empirical evaluation of the performance bias of ear recognition techniques with respect to the gender of subjects shown in the ear images.
\end{itemize}

\section{Related work}\label{Sec: Related}

In this section we present relevant prior work. We first discuss existing datasets of ear images captured in unconstrained conditions  and then elaborate on recent research on ear recognition in the wild.

\textbf{Datasets for unconstrained ear recognition.} 
Numerous datasets are currently available for research in ear recognition~\cite{Survey2017}. These include early datasets, such as UND image collections E, F, G and J2\footnote{Available from: \url{https://cvrl.nd.edu/projects/data/}}, the USTB~\cite{USTBdatasets} datasets or the popular IITD dataset from~\cite{Kumar2012}, but also newer ones, such as AWE~\cite{Survey2017}, AWEx~\cite{ncaa}, WebEars~\cite{zhang2017ear}, or Helloear~\cite{zhangUSTBHelloearLargeDatabase2017}. 

For a considerable period of time, research on ear recognition relied on the (early) controlled datasets listed above, where images were typically captured in constrained conditions, under full profile view, limited illumination and occlusion variations and at high-resolution. Furthermore, the same acquisition hardware was commonly used for the image collection process. To advance the field and provide the research community with a more challenging problem, the WPUT~\cite{Frejlichowski2010} dataset was introduced in 2010. The dataset offered images with a significantly wider range of appearance variability (across pose, illumination and occlusions), but was still recorded in controlled, laboratory-like settings. While being more representative of real-world conditions than earlier datasets, WPUT was not widely adopted, likely due to the low recognition rates achieved by the techniques being proposed in the literature. 

Recent years have seen the introduction of completely unconstrained ear datasets, where images were not  captured specifically for research purposes, but were instead compiled from pre-recorded data collected from the web. The first such dataset was the AWE dataset described by Emer\v{s}i\v{c} et al. in~\cite{Survey2017,pedced} and it's later extension AWEx~\cite{ncaa}. Zhang et al.~\cite{zhang2017ear} presented another dataset caled WebEars with similar characteristics to AWE, but with a focus on ear detection. Zhou and Zaferiou~\cite{Ref2} introduced the In-the-wild Ear Dataset, which was constructed from an existing unconstrained web-based dataset of face images. In a more recent collection effort, Zhang et al.~\cite{zhangEarVerificationUncontrolled2018,zhangUSTBHelloearLargeDatabase2017,zhang3DEarNormalization2017} collected a dataset of ear video sequences recorded with a mobile phone.

The dataset used for UERC 2019 follows the trends outlined above and consists of images collected from the web. However, different from competing datasets, the UERC data contains several thousand of images with more than $3,000$ identities, which makes it the largest dataset of unconstrained ear images publicly available. Given the results of UERC 2017, where the dataset was already used, it also represent one of the most challenging image collections in the field of ear recognition. 
\begin{figure}[!tb!]
\begin{center}
	\includegraphics[width=0.9\columnwidth]{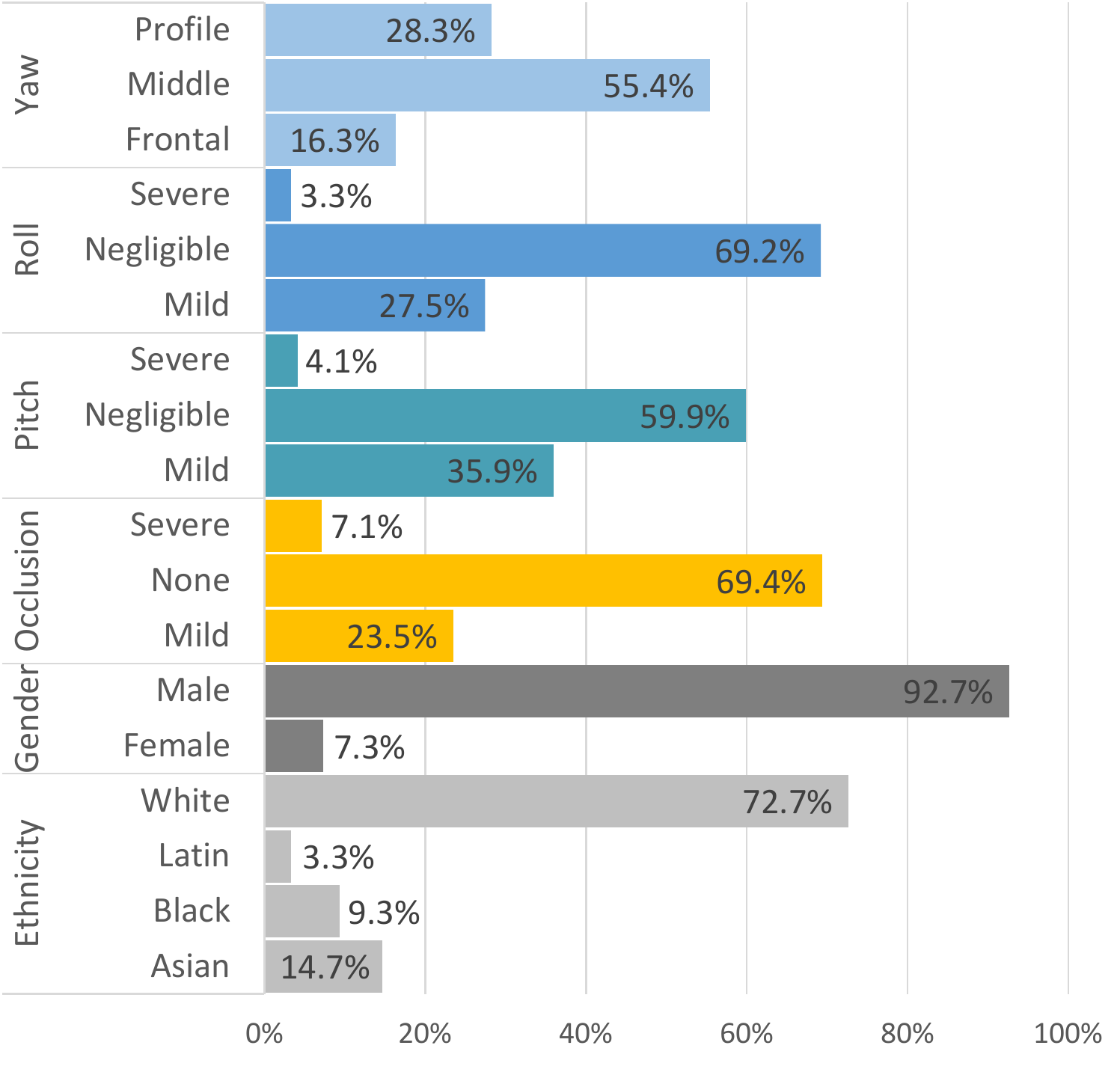}\vspace{-2mm}
\caption{Distribution of the available annotations in the AWEx dataset. Each image is annotated with multiple labels that define: the yaw, roll and pitch angles, the amount of occlusion, the gender and the ethnicity of the subjects. These labels allow for a fine-grained analysis of the recognition results. The figure is best viewed in color.  \vspace{-5mm}
}
\label{barGraphs}
\end{center}\end{figure}
\begin{figure}[tb] \begin{center}
%
	\includegraphics[width=1\columnwidth]{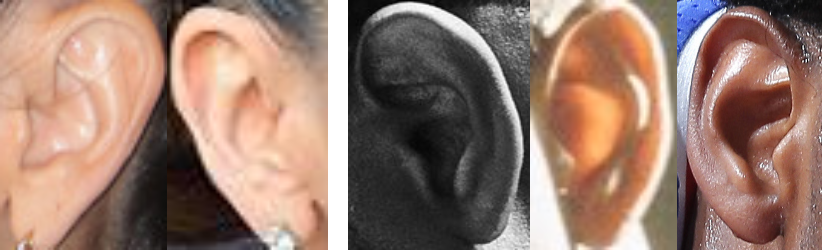}
\caption{Sample images of two subjects from the sequestered dataset. Note the large appearance variability. 
The sequestered data is used in UERC to assess the generalization abilities of ear recognition techniques. \vspace{-3mm}}
\label{sequesteredDataset}
\end{center}\end{figure}

\textbf{Ear recognition in the wild.} With the introduction of unconstrained ear datasets (described above), techniques for ear recognition started to shift away from the then dominant descriptor-based methods and techniques combining descriptors with subspace projection techniques (see~\cite{Survey2017,Pflug2014a} for details) to more powerful approaches that could better capture the considerable appearance variability of ear image data when recorded in uncontrolled real-life conditions. Specifically popular in this regard are methods are relying on convolutional neural networks (CNNs), as evidenced by the significant body of work adopting these models for ear recognition, e.g.,~\cite{emervsivc2017unconstrained, ziga@bwild,eyiokurDomainAdaptationEar2017,ncaa,dodgeUnconstrainedEarRecognition2018,zhangEarVerificationUncontrolled2018,hansleyEmployingFusionLearned2018,omaraLearningPairwiseSVM2018,almisrebUtilizingAlexNetDeep2018,yingHumanEarRecognition2018}.  

With UERC 2019, we study many of the approaches referenced above and conduct one of the first independent evaluations of ear recognition techniques so far. We assess all techniques on the same datasets and protocol and, consequently, contribute to a better understanding of the current capabilities and limitations of ear recognition technology.   
\section{Methodology}\label{Sec: methods}

In this section, we present the methodology adopted for UERC 2019. We first discuss the  datasets used in the challenge and then describe the experimental protocol, performance metrics, and starter kit distributed to the challenge participants. We conclude the section with a summary of  the submitted approaches. 

\subsection{UERC 2019 datasets}\label{SSec: dataset}

Two sets of image data were used for UERC 2019. The first (the \textit{public dataset} hereafter) was distributed to all participants and featured training and testing data for the first stage of the challenge. The second dataset was a hold-out dataset (the \textit{sequestered dataset} hereafter) used in the second stage of the challenge by the UERC organizers to evaluate the generalization capabilities of the best performing algorithms. This dataset was not available to the participants during development and contained data with slightly different characteristics than the public dataset. A detailed description of both datasets is given below.

\begin{table*}[!tb]
\renewcommand{\arraystretch}{1.2}
\caption{Summary of the UERC data splits and experimental setup. A public dataset used for development and testing was used in the first stage and a sequestered dataset was used in the second stage to test the generalization abilities of the participating approaches to data not seen during development. \vspace{-1mm}
}
\label{Tab: experimental protocol}
\centering
\footnotesize
\setlength\tabcolsep{5.5pt}
\begin{tabular}{ll lcccc}
\hline \hline
UERC 2019 Datasets & Data splits 				& 	Specification 			& \#  Images 	& \# Subjects 	& \# Images per subject 	& Total \# Images (\# Subjects)\\
 \hline 
\multirow{ 3}{*}{Public dataset} & {Training split}		&   From AWEx 	& $1,500$    & $150$			& 	$10$			&  $1,500 \text{ }(150)$\\
\cline{2-7}                                     
& \multirow{ 2}{*}{Testing split}		& 	From AWEx 		& $1,800$ 	& $180$			& 	$10$				 	& \multirow{ 2}{*}{$9,500 \text{ }(3,540)$}\\
	&								&   From UERC 2017			& $7,700$ 	& $3,360$		& 	Variable 			& \\ \hline
Sequestered dataset & Testing split & UERC 2019 specific & 500 &  50 & 10 & 500 (50) \\
\hline 
\hline
\end{tabular}
\end{table*}

\textbf{The public UERC dataset} contained  $11,000$ ear images of $3,690$ subjects. The main part of this data was taken from the Extended Annotated Web Ears (AWEx)~\cite{emervsivc2017unconstrained,Survey2017} dataset and comprised $3,300$ ear images of $330$ subjects - $10$ images per subject. Images of the dataset were collected from the web and exhibit considerable appearance variations not usually seen with other ear datasets. 
AWEx images come with a rich set of annotations, such as head rotations in different  directions, gender and race labels, labels for the presence of occlusions and are, therefore, used as the basis for most of analysis presented in Section~\ref{Sec: exper}. A summary of the available annotations and their distribution over the AWEx images is shown in Figure~\ref{barGraphs}. The reader is referred to~\cite{ncaa} for more information on the dataset.
\begin{table*}[!tb]
\renewcommand{\arraystretch}{1.1}
\caption{High-level comparison of submitted approaches. The submissions were dominated by deep-learning models and hybrid approaches combining learned and hand-crafted features. Except for three submissions (i.e., PHOG, IFLBP and BIOR) and the two baselines, all participating approaches used multiple descriptors to represent ear images. More detailed descriptions of the submitted approaches are given in the Appendix.\vspace{-1mm}} 
\label{tab: approach summary}
\centering
\footnotesize
\resizebox{\textwidth}{!}{%
\begin{tabular}{ l  l lrrrrrr}
\hline \hline
Approach   & Participant$^\star$ & Approach summary \& Relevant reference            &     Basic features & HC/LR$^{\dagger}$  & Feature dim. & MS$^{\star\star}$ & Loc./Hol. & \# LP$^{\ddagger}$\\ \hline

PHOG	&NMAM IT \& JSS STU	& Pyramid of Histograms of Oriented Gradients; CosS$^{1}$~\cite{bosch2007representing} 	&			HOG	&	HC	&	$5,096$	&	0	&	Local	&	0	\\
IFLBP	& NMAM IT \& JSS STU	&Intuitionistic Fuzzy Local Binary Pattern; CosS~\cite{ansari2016feature} 	&	   	LBP	&	HC	&	$256$	&	0	&	Local	&	0	\\
SEP	& NASK \& WUT	& Siamese ResNet-50; CosS~\cite{he2016deep}	&			CNN	&	LR	&	$2,048$	&	$94$	&	Holistic	&	$26$M	\\
CH-Fus	&USF \& UFBA	&
Score-level fusion of CNN and HOG feat.; CosS, ChiS$^{2}$; SUM$^5$~\cite{hansleyEmployingFusionLearned2018}	&			HOG; CNN	&	Both	&	$9,224$	&	$146$	&	Both	&	$38$M	\\
CHP-Fus	& USF \& UFBA	& Score-level fusion of $6$ feat. types; CosS, ChiS; SUM~\cite{hansleyEmployingFusionLearned2018,ncaa}	&			HOG; POEM; CNN$^6$	&	Both	&	$20,552$	&	$226$	&	Both	&	$59$M	\\
MLE-CNN	& KIST	&  Ensemble of $4$ CNN models: $2\times$ VGG-16+ResNet-50; CosS; SUM~\cite{parkhi2015deep,he2016deep}	&			CNN	&	LR	&	$8,144$	&	$1,270$	&	Holistic	&	$327$M	\\
SiamCNN	& IITB	& Siamese twin Resnet-18; EucS$^{3}$	&			CNN	&	LR	&	$2,304$	&	$12$	&	Holistic	&	$12$M	\\
ScNet-5	& ITU$^{*}$	& Deep Cascaded Score Level Fusion (fusion learning; $41$ experts)~\cite{kacarScoreNetDeepCascade2018}	&			Multiple$^7$	& LR &	cca. $12$k-$30$k	&	$15,300$	&	Both	&	$2$G	\\
VGGEar	& USTB	& Ensemble of $3$ VGG models; CosS~\cite{parkhi2015deep}	&			CNN	&	LR	&	$166$	&	$589$	&	Holistic	&	$120$M	\\
LC-Fus	& ITU$^{**}$	& Score-level fusion of CNN and LBP features; CosS; SUM~\cite{parkhi2015deep,emervsivc2017unconstrained} 	&			LBP; CNN	&	Both	&	$14,067$	&	$515$	&	Both	&	$120$M	\\ 
BIOR		& NITR    & Patch-based wavelet energy features; CanS$^{4}~\cite{chowdhury2018applicability}$ 			&     Wavelets  				&    HC		&  $21,096$	    	&    0      & Local     & 0 \\ 
\hline
LBP-Base          & Organizers   & Local Binary Patters; CosS~\cite{pietikainen2011local,emervsivc2017unconstrained}        &  LBP& HC                 & n/a           & 0        & Local    & 0\\
VGG-Base          & Organizers   & VGG-16; CosS~\cite{parkhi2015deep,emervsivc2017unconstrained}        &  CNN-based & LR                 & $4,096$        & $600$  & Holistic &   $120$ M \\

\hline \hline
\multicolumn{6}{l}{ $^{*}$Department of Electronics and Communication Engineering,\hspace{5mm} $^{**}$Department of Computer Engineering} \\
 \multicolumn{9}{l}{$^{\star}$Refer to author affiliations on first page for the abbreviations, \hspace{4mm}$^{\star\star}$MS - Model size in MB, \hspace{4mm} $^{\dagger}$HC - hand-crafted, LR - learned, \hspace{4mm}$^{\ddagger}$\# LP - Number of learnable parameters}\\
  \multicolumn{9}{l}{$^{1}$CosS - cosine scoring, \hspace{4mm} $^{2}$Chi-squared scoring, \hspace{4mm} $^3$Euclidian-distance scoring, \hspace{4mm} $^{4}$Canberra-distance scoring, \hspace{4mm} $^5$(Weighted) SUM-rule fusion}\\
  \multicolumn{9}{l}{$^{6}$ An their modified versions, \hspace{1mm}$^{7}$ HOG, Gabor, LBP, LPQ, BSIF, 10 CNNs (AlexNet, VGG-16, VGG-19, GoogLeNet, ResNet-101, ResNet-18, SqueezeNet, InceptionV3, IncetionResNetV2 and DenseNet) }
\end{tabular}
}
\end{table*}

The rest of the data in the public dataset was taken from the UERC 2017 dataset. Images from this part were also gathered from the internet to ensure characteristics and variability similar to the AWEx images. For the data collection procedure, a pool of candidate face images was first collected. An automatic post-processing procedure involving the ear detection approach from~\cite{pedced} was then used to identify potential ear regions. The detected (ear-candidate) regions were manually inspected to check for the quality of the detections and false positives and poorly localized ear regions were excluded. All detected ear images were tightly cropped and left as they were. No additional effort was made to align the images to a common side - right or left. Different from the AWEx data, these additional images exhibited large variability in size - the smallest images containing only a few hundred pixels, whereas the largest having close to $400$k pixels. The average pixel count per image was $3,682$. Sample images from the public dataset are shown on the right side of Figure~\ref{datasetComparison}.

\textbf{The sequestered UERC dataset } consisted of $500$ images of $50$ subjects. This dataset was intentionally kept small to allow for quick testing of the submitted models. Similarly to the public dataset, this dataset featured images captured in the wild, but the data collection procedure was conducted with the help of student volunteers. The students were asked to find ear images of arbitrary subjects online, so a certain level of human-bias was introduced into dataset, as images were typically of somehow better resolution and quality and of slightly different visual characteristics compared to the public dataset. To make sure there was no overlap in terms of identities with the public part of the UERC data, the students checked the identity of their subjects against the subject list of the public dataset. Example images from the sequestered datasets are shown in Figure~\ref{sequesteredDataset}.

\subsection{Experimental setup}\label{SSec: protocol}

UERC was conducted in two separate stages. The first stage focused on model development and testing using the public UERC dataset, whereas the second stage addressed model generalization on the sequestered dataset, i.e., with image data not seen during development.

For the first stage, the public UERC dataset was partitioned into disjoint training and testing splits. The training split consisted of $2,304$ images of $166$ subjects from the AWEx dataset, whereas the testing split contained the remaining AWEx data and the rest of the images from UERC 2017 - see Table~\ref{Tab: experimental protocol} for details. The training split was used to train feature extraction and classification models (e.g., CNNs, classifiers, etc.) and set potential hyper-parameters, while the testing split was reserved exclusively for the  performance evaluation. Using images from the testing split during development and training was not allowed. To ensure that the participants followed the experimental protocol, a small amount of controlled label noise was introduced into the testing split of the dataset. Results for the mislabeled data were then checked manually by the organizers. 

Participants were asked to submit results in the form of a similarity matrix of size $7,742 \times 9,500$ to the organizers. This similarity matrix served as the basis for the evaluation and allowed the organizers to analyze different aspects of the submitted recognition approaches. The similarity matrix was generated by comparing $7,742$ probe images (corresponding to $1,482$ subjects) to all $9,500$ gallery images (corresponding to $3,540$ subjects) in an \textit{all-vs-all} experimental protocol using a selected similarity measure or matching approach. Here, the entire testing split of the public dataset was used for the gallery, while the $7,742$ probe images represented a subset of the $9,500$ gallery images, featuring only subjects with at least $2$ images in the dataset. Each similarity score had to be generated based solely on the comparison of two ear images and no information about other subjects in the testing split of the UERC data was allowed to be used for the score generation. 

In the second stage of UERC 2019, the top performing algorithms from the first stage were tested on a sequestered dataset. For this part, the sequestered data was anonymized (i.e., image names were randomized) and distributed among participants. The participants then had $2$ weeks to compute similarity scores for a number of predefined comparisons and return a second $500\times500$ similarity matrix for scoring. It was not allowed to make any changes to the submitted algorithms during this stage, and the submitted approaches had to be tested with the same configuration and parameter settings as in the first UERC stage.

\subsection{Performance metrics}

All submitted algorithms were tested for identification accuracy and Cumulative Match Score (CMC) curves were generated to visualize results. To report performance, the following quantitative metrics were used: \textit{i)} the recognition rate at rank one (rank-$1$), which corresponds to the fraction of probe images, for which an image of the correct identity was retrieved from the gallery as the top match, \textit{ii)} the recognition rate at rank five (rank-$5$), which corresponds to the fraction of probe images, for which an image of the correct identity was among the top five matches retrieved from the gallery, and \textit{iii)} the area under the CMC curve (AUC), which 
measured the performance of the recognition approached at different ranks. The latter was computed by normalizing the maximum  rank (i.e., the number of distinct gallery identities) of the experiments to $1$. Rank-$1$ scores were used to determine the final ranking of the submitted algorithms.



\subsection{The UERC 2019 starter kit}

A MATLAB starter kit was provided to the challenge participants to help them get started with the challenge. The starter kit included implementations of two baseline algorithms, i.e., \textit{i)} an LBP-based approach (LBP-Base hereafter) and \textit{ii)} a CNN-based model build around the VGG-16 architecture (VGG-Base\footnote{See Table~\ref{tab: approach summary} and Appendix for details.} hereafter) as well as a sample script that implemented the predefined experiments protocol for the two baselines. This sample script ensured that participants were able to easily generate compliant similarity matrices for scoring. Additionally, scripts and functions for assessing the sensitivity of the recognition models to various data characteristics were also distributed. The starter kit was put online and will stay available after the challenge to help researchers to quickly analyze their models on the UERC datasets.  Please refer to {\small \url{http://ears.fri.uni-lj.si/uerc19}} for more information.

\subsection{Participating approaches}


The challenge received a total of $13$ submissions from research groups belonging to $12$ institutions. As noted in the previous section, $2$ of these submissions represented baseline models made available by the UERC 2019 organizers.  The submissions featured algorithms with different characteristics, model sizes, and different numbers of learnable parameters, as summarized in Table~\ref{tab: approach summary}. In line with recent trends, most of the submissions focused on deep learning, with $5$ approaches (or $38\%$) using solely CNN models for image representation, i.e., SEP, MLE-CNN, SiamCNN, VGGEar and VGG-Base, and $4$ submissions (or $31\%$) relying on a hybrid combination of hand-crafted and learned image descriptors, i.e., CH-Fus, CHP-Fus, ScNet-5, and LC-Fus. Only $4$ techniques (or $31\%$) used  hand-crafted features exclusivelly, i.e., PHOG, IFLBP, BIOR and LBP-Base.

Despite the fact that the majority of submissions involved deep learning models ($69\%$), these models differ significantly in number of parameters that need to be learned during training. The smallest models, i.e., SEP, CH-Fus, CHP-Fus and SiamCNN all have less than $60$ million parameters to train. The somewhat larger models, i.e., VGGEar, LC-Fus, VGG-Base, need to learn around $120$ million parameters, whereas the ensemble MLE-CNN approach requires optimizing for more than $300$ million learnable parameters. The largest model by a significant margin is ScNet-5 with $2$ billion parameters. 

Interestingly, many of the submitted approaches ($6$ submissions or $46\%$) relied on multiple image representations by either \textit{i)} fusing different descriptors (CH-Fus, CHP-Fus, LC-Fus), \textit{ii)} using ensembles of CNN models (MLE-CNN, VGG-Ear), or \textit{iii)} learning to fuse multiple ear experts (ScNet-5). As we show in the experimental section, these approaches are among the best performing approaches in the challenge, which suggests that a single representation is not sufficient to capture the complex appearance variations seen with unconstrained ear images.   

The reader is referred to the Appendix for a more complete description of the participating approaches.
\begin{figure}[tb]
  \centering
  \includegraphics[width=1\columnwidth]{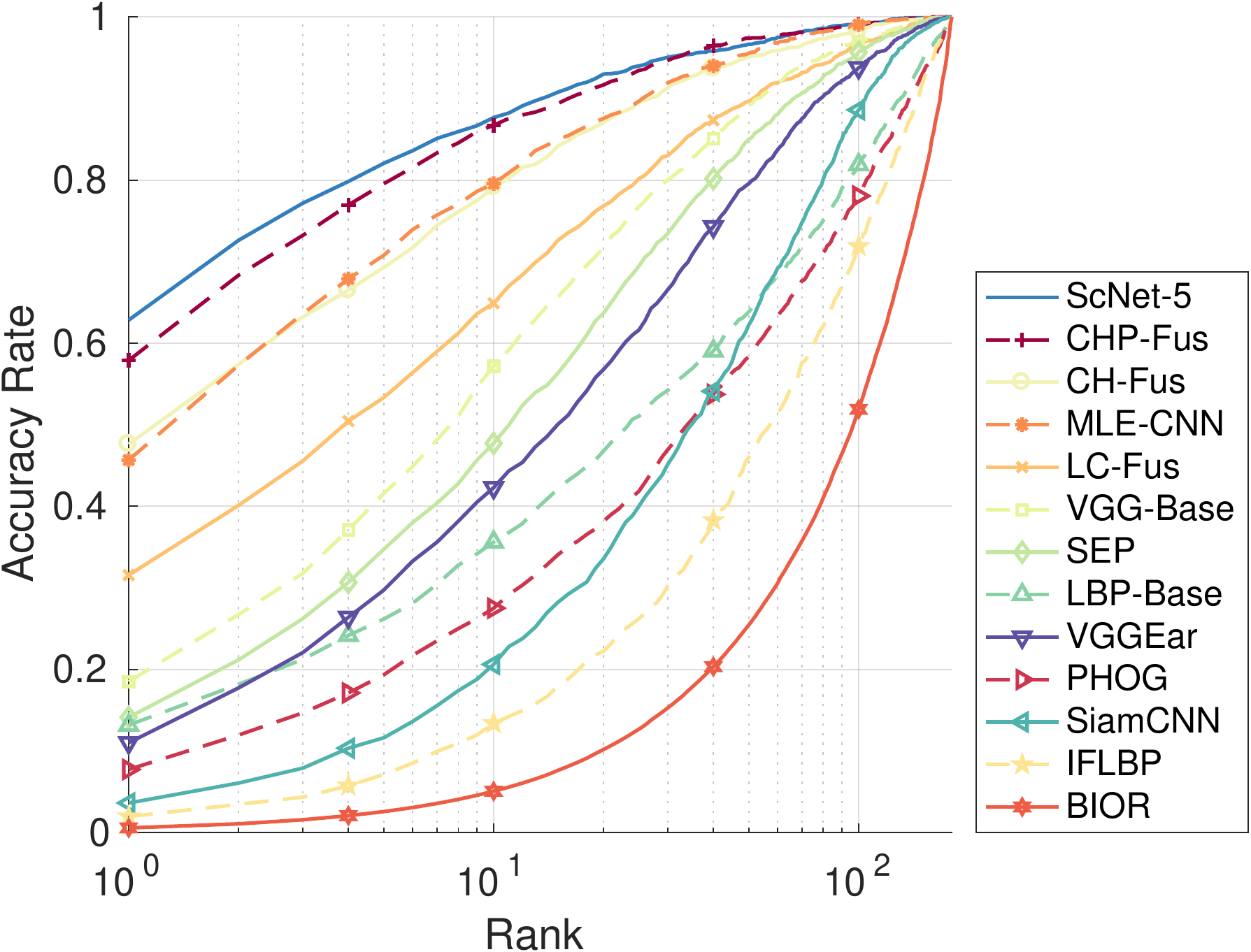}
\caption{CMC curves of the submitted approaches generated on the testing split of the AWEx dataset (shown in logarithmic scale). The legend on the right is sorted according to the rank-1 performance. Best viewed in color.}
\label{fig:aweOnly}
\end{figure}


\section{Experiments and results}\label{Sec: exper}

In this section we present the results of the UERC 2019. We conduct an in-depth analysis of all the submitted approaches and use the submitted results to study different aspects of existing ear recognition technology, such as sensitivity to ear rotations (in terms of yaw, tilt and roll angles) and occlusions, performance over different image resolutions, and potential bias with respect to gender.

\subsection{Comparative assessment}

\begin{figure*}[!tb]
\centering
{\captionsetup{justification=centering}
  \begin{subfigure}[b]{0.32\textwidth}
    \includegraphics[width=\columnwidth]{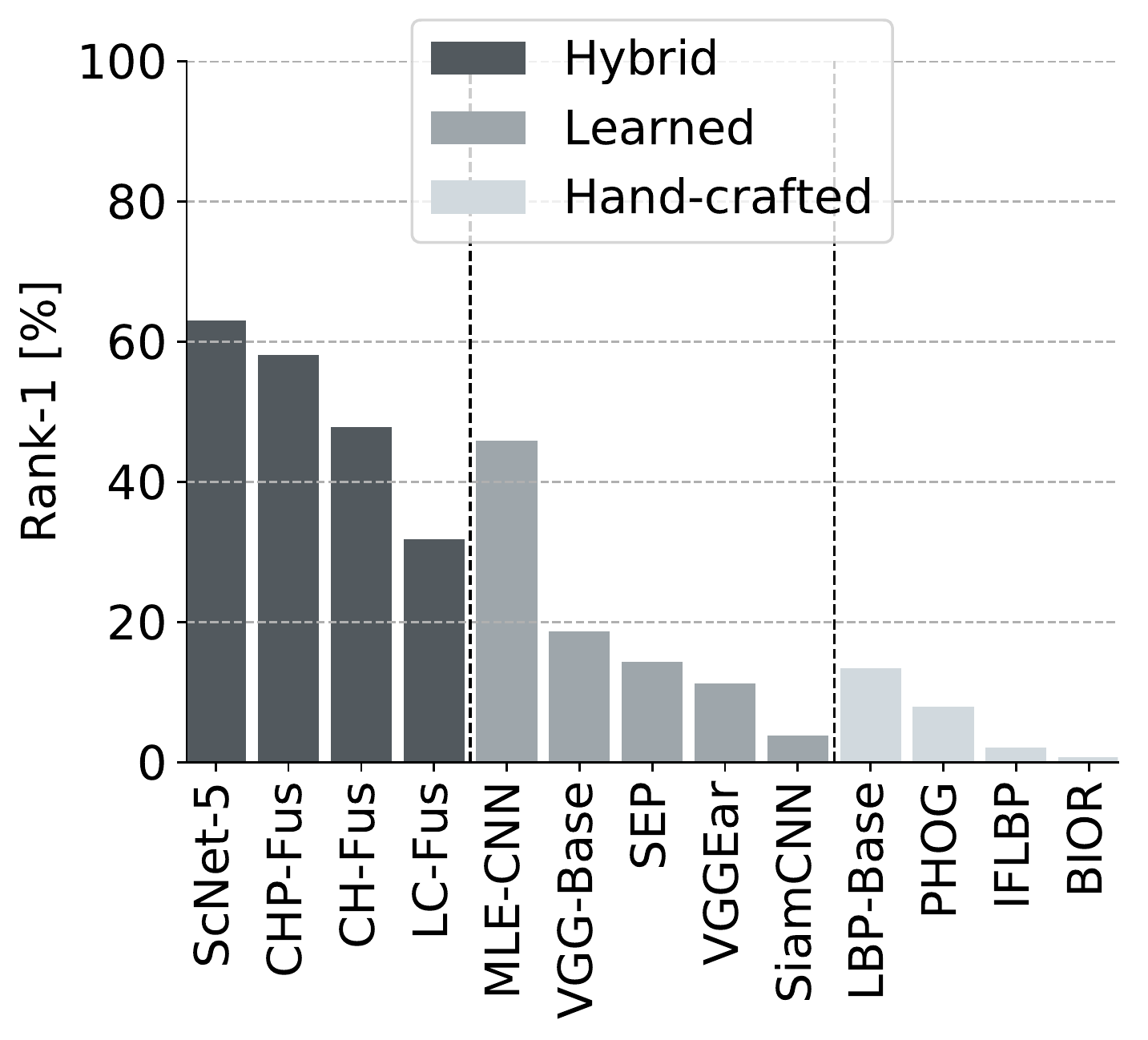}\vspace{-1mm}
    \caption{Comparison by feature type (on AWEx)}
    \label{fig:HCvsLR}\vspace{4mm}
  \end{subfigure}
  \hfill
  \begin{subfigure}[b]{0.32\textwidth}
    \includegraphics[width=1\columnwidth]{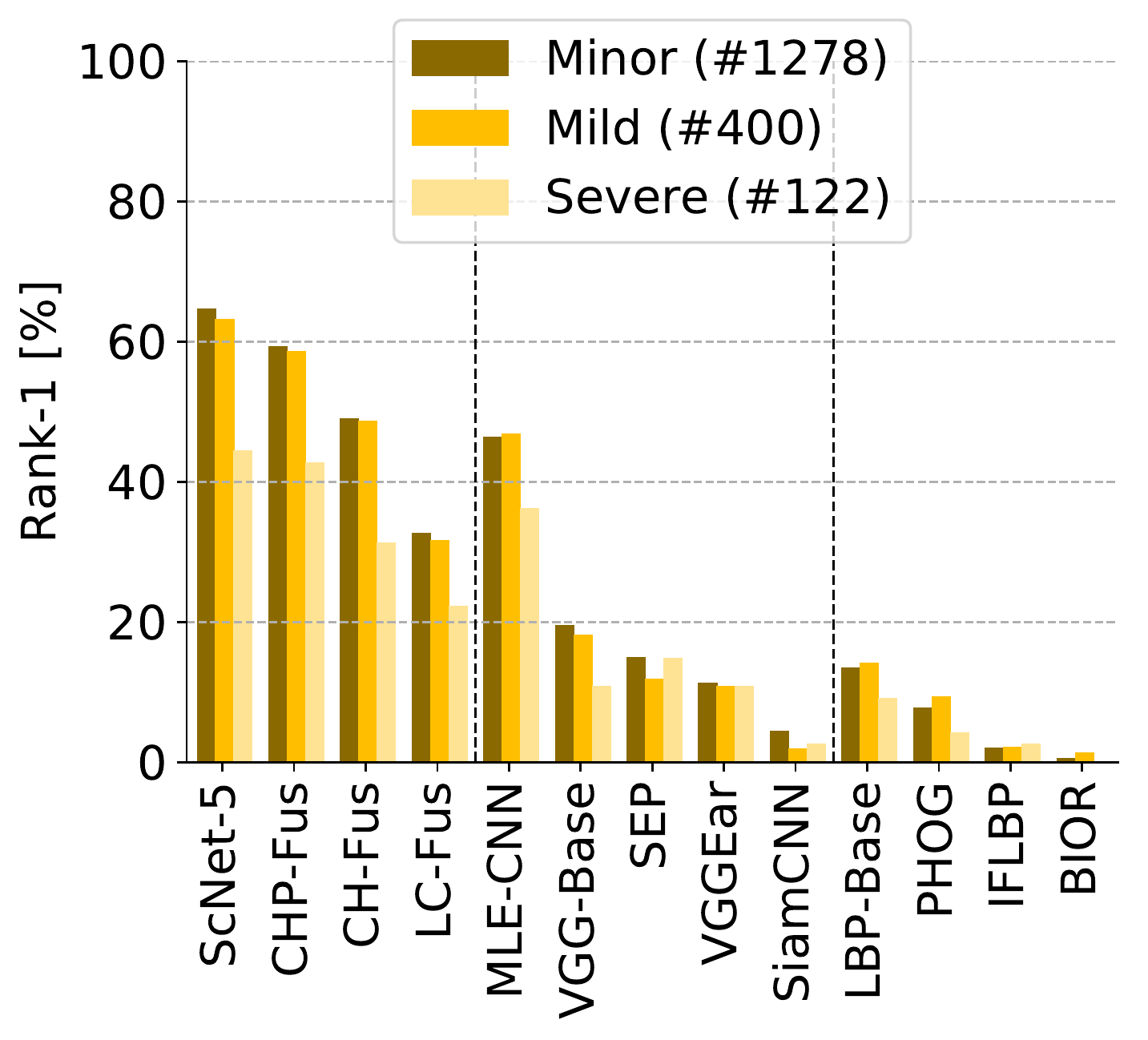}\vspace{-1mm}
    \caption{Impact of occlusion (on AWEx)}
    \label{fig:covOcclusion}\vspace{4mm}
    \end{subfigure}
  \hfill
  \begin{subfigure}[b]{0.32\textwidth}
    \includegraphics[width=1\columnwidth]{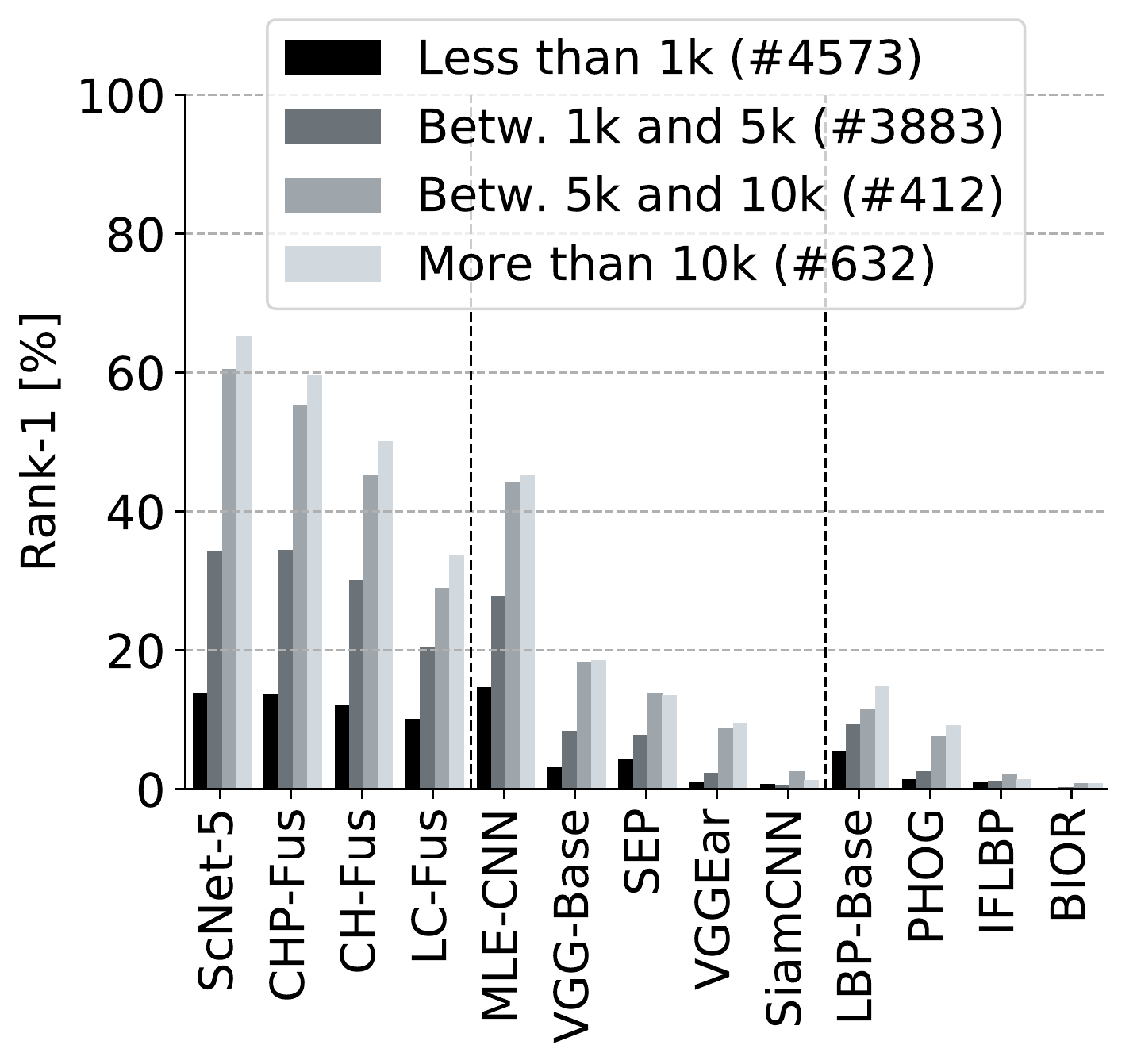}\vspace{-1mm}
    \caption{Impact of image resolution (on UERC)}
    \label{fig:covResolution}\vspace{4mm}
  \end{subfigure}
\newline 
  \begin{subfigure}[b]{0.32\textwidth}
    \includegraphics[width=\textwidth]{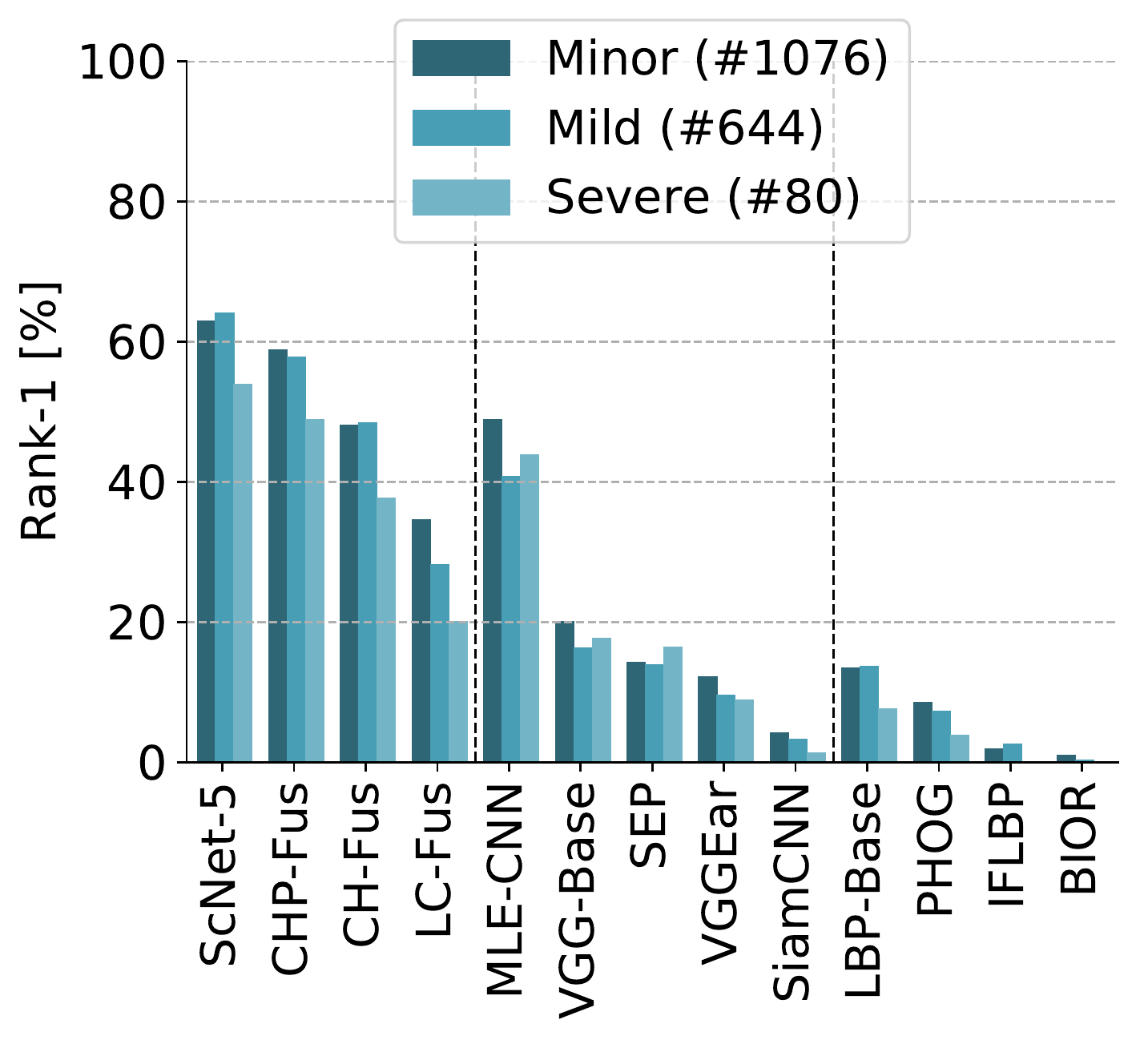}
    \caption{Impact of rotation - pitch (on AWEx)}
    \label{fig:covPitch}
  \end{subfigure}
  \hfill
  \begin{subfigure}[b]{0.32\textwidth}
    \includegraphics[width=\textwidth]{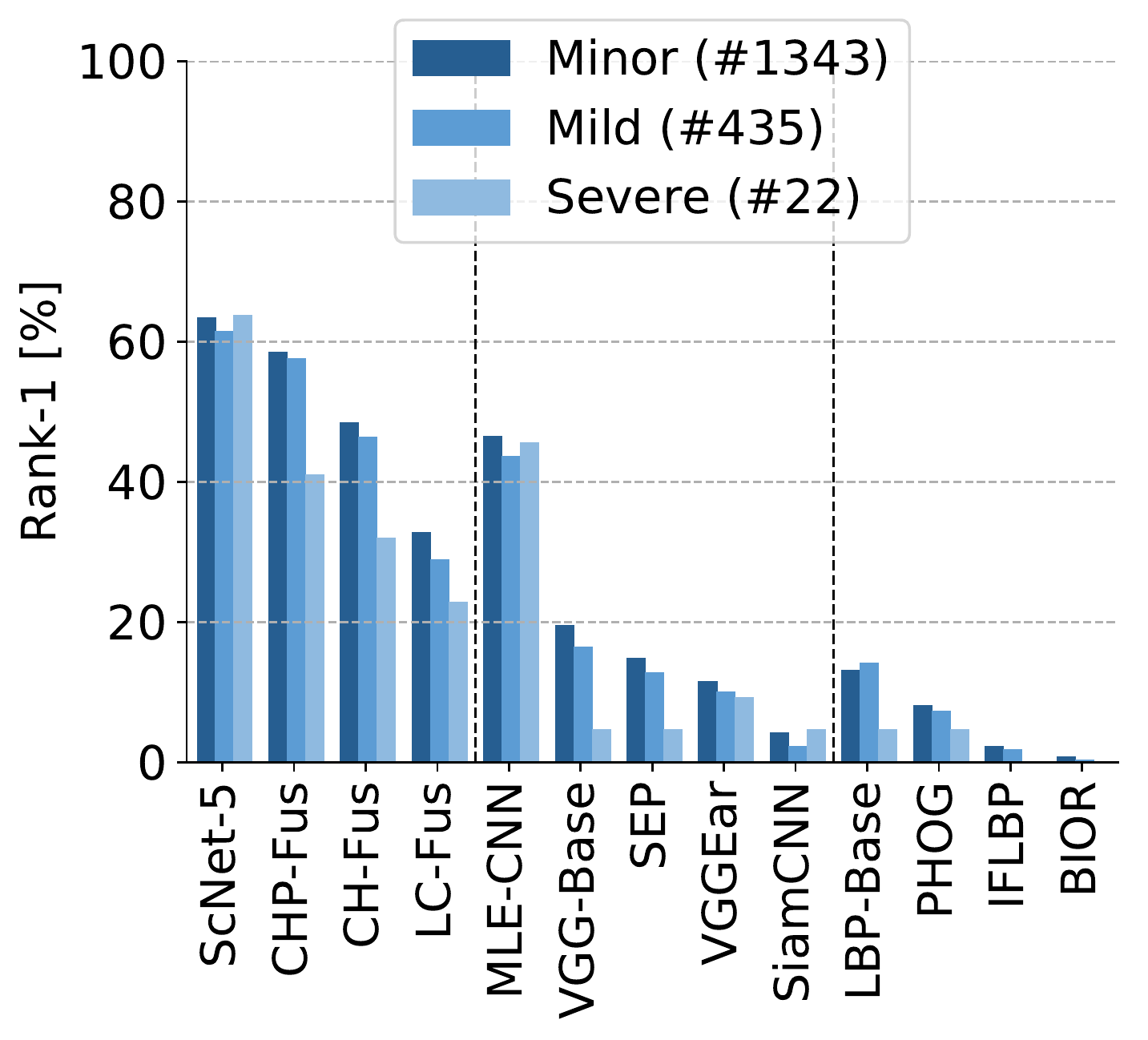}
    \caption{Impact of rotation - roll (on AWEx)}
    \label{fig:covRoll}
  \end{subfigure}
  \hfill
  \begin{subfigure}[b]{0.32\textwidth}
    \includegraphics[width=\textwidth]{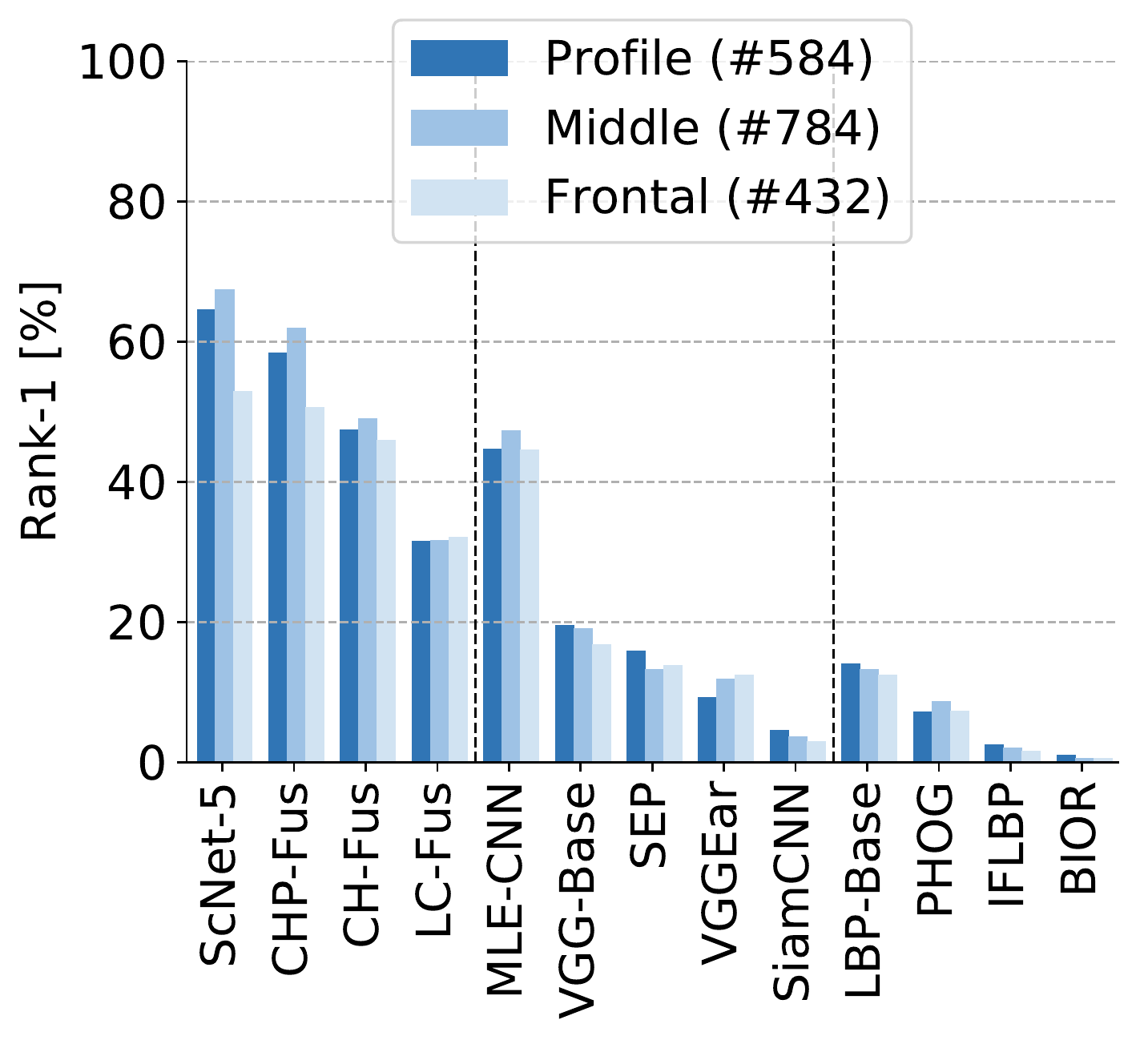}
    \caption{Impact of rotation - yaw (on AWEx)}
    \label{fig:covYaw}
  \end{subfigure}
 } 
  \caption{Sensitivity to various covariates. The bar-graphs show: (a) baseline rank-1 recognition rates on the AWEx test set for all tested methods, (b) impact of different occlusion levels, (c) impact of resolution changes ``k'' denotes $1,000$ pixels), (d)-(f) impact of head rotation in terms of pitch, roll and yaw angles. The numbers within the brackets indicate the number of images for each label. All the approaches were first grouped by  type (hand-crafted, learned or hybrid) and then sorted according to rank-1 performance on all images. The vertical delimiters separate feature-type groups. Best viewed in color.}
  \label{fig:cov}
\end{figure*}

\textbf{Experiments on AWEx.} In our first series of experiments, we compare the submitted approaches on the testing split of the AWEx dataset (involving $180$ subjects), as described in Section~\ref{Sec: methods}. The results in Figure~\ref{fig:aweOnly} and Table~\ref{tab:aweOnly} 
show that hybrid techniques that combine hand-crafted and learned features achieve the best performance, followed by methods relying solely on learned image descriptors. Submissions using only hand-crafted descriptors are less competitive and exhibit a considerable performance gap with respect to approaches that incorporate deep-learning models. These trends are clearly visible in Figure~\ref{fig:HCvsLR}, where a bar graph with rank-1 recognition rates is presented for the three different groups of methods. 

The overall top performed in this series of experiments with a rank-1 recognition rate of $62.8\%$ is ScNet-5, followed closely by CHP-Fus with a rank-1 score of $57.9\%$. This comparison is particularly interesting when also considering the model size, where ScNet-5 is the biggest and most resource hungry approach with a model size of around $15$ GB, while the CHP-Fus model uses a model of only $226$ MB in size. Ranked third and fourth are the CH-Fus and MLE-CNN models, which achieve similar performance with rank-1 values of around $45\%$. The next model, LC-Fus, is again a hybrid approach and achieves a rank-1 recognition rate of $31.6\%$. The SEP, VGGEar, PHOG, SiamCNN, IFLBP and BIOR approaches perform similarly or below the baselines provided by the organizers.


\begin{table}
\caption{Comparative evaluation on the testing split of the AWEx dataset (involving 180 subjects). The results are ordered according to rank-1 scores.}
\label{tab:aweOnly}
\centering
\footnotesize
\setlength\tabcolsep{5.5pt}
\begin{tabular}{ll rrr}
\hline
Approach & HC/LR$^{\dagger}$ & Rank-1 [\%] & Rank-5 [\%] & AUC \\
\hline
\rowcolor{claC} ScNet-5   & Both   & $62.8$ & $82.1$ & $0.966$\\
\rowcolor{claC} CHP-Fus   & Both   & $57.9$ & $79.6$ & $0.964$\\
\rowcolor{claC} CH-Fus    & Both   & $47.6$ & $69.3$ & $0.946$\\
\rowcolor{claB} MLE-CNN   & LR     & $45.7$ & $70.8$ & $0.951$\\
\rowcolor{claC} LC-Fus    & Both   & $31.6$ & $53.3$ & $0.907$\\
\rowcolor{claB} SEP       & LR     & $14.1$ & $34.7$ & $0.867$\\
\rowcolor{claB} VGGEar    & LR     & $11.0$ & $29.7$ & $0.839$\\
\rowcolor{claA} PHOG      & HC     & $7.7 $ & $19.3$ & $0.703$\\
\rowcolor{claB} SiamCNN   & LR     & $3.6 $ & $11.6$ & $0.746$\\
\rowcolor{claA} IFLBP     & HC     & $1.9 $ & $7.1 $ & $0.624$\\
\rowcolor{claA} BIOR      & HC     & $0.6 $ & $2.6 $ & $0.474$\\
\hline              
\rowcolor{claB} VGG-Base  & LR     & $18.5$ & $41.6$ & $0.895$\\
\rowcolor{claA} LBP-Base  & HC     & $13.2$ & $26.2$ & $0.743$\\
\hline                                       

\multicolumn{5}{l}{$^{\dagger}$\footnotesize \colorbox{claA}{HC - hand-crafted,} \colorbox{claB}{LR - learned,} \colorbox{claC}{Both (hybrid - LR/HC)}}

\end{tabular}\vspace{-2mm}
\end{table}

\begin{figure}[tb]
  \centering
  \includegraphics[width=1\columnwidth]{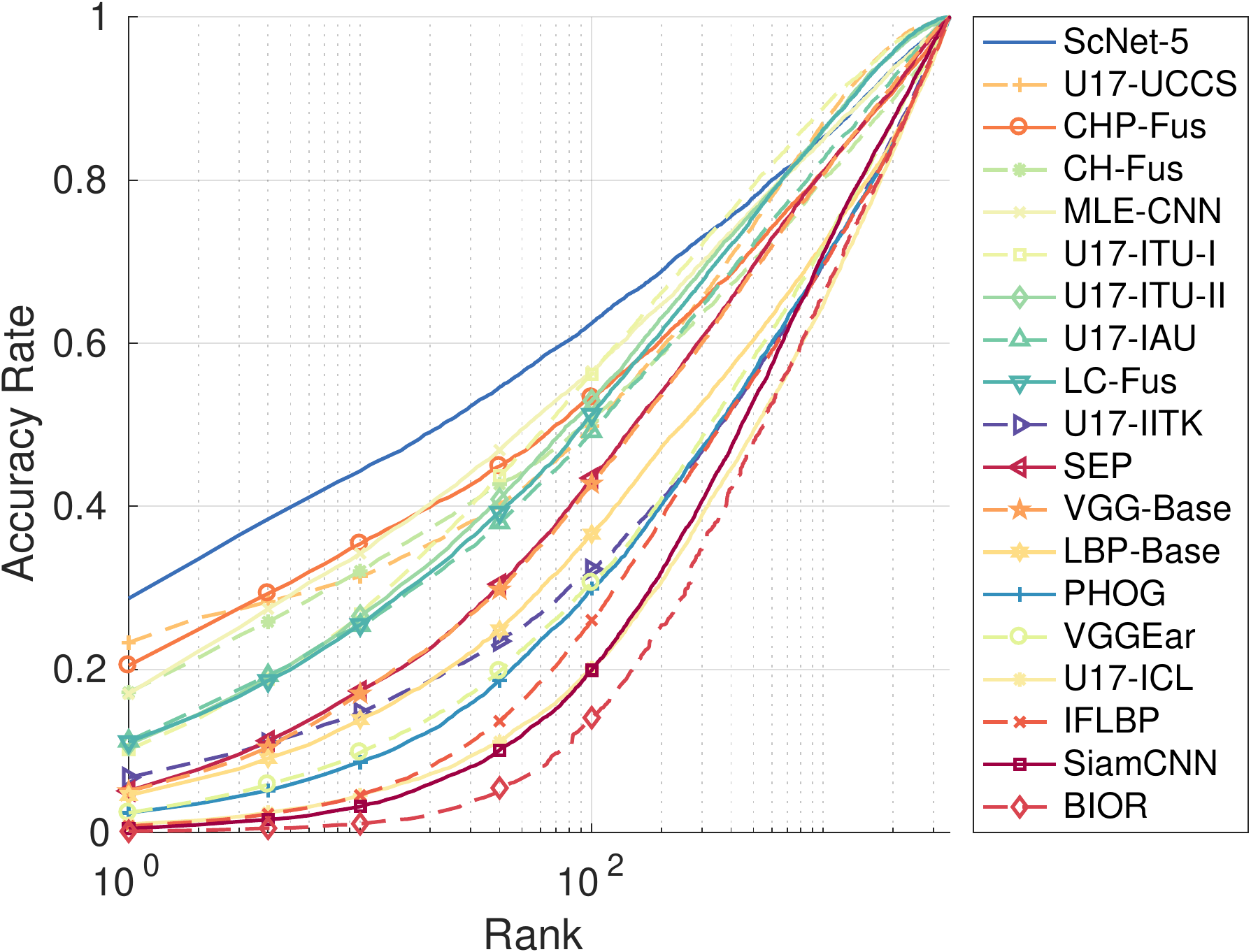}\vspace{-2mm}
\caption{CMC curves of the submitted approaches generated on the complete UERC test dataset involving $3,540$ subjects (shown in logarithmic scale). The legend is sorted according to the rank-1 scores. U17 denotes approaches from UERC 2017. Best viewed electronically and in color.}
\label{fig:scale}
\end{figure}

\textbf{Experiments on the UERC dataset.} In the second series of experiments, we compare the submitted approaches on the testing split of the entire UERC dataset. As suggested in Section~\ref{Sec: methods}, these experiments involve  $9,500$ images and $3,540$ subjects and aim at evaluating the scalability of the recognition techniques. To facilitate comparisons with submissions from UERC 2017~\cite{emervsivc2017unconstrained}, we also report results for the techniques participating in the first ear recognition challenge (marked U17) in Figure~\ref{fig:scale} and Table~\ref{tab:scale}. Note that due to a minor error in the label file used for UERC 2017, the reported results were recomputed and differ slightly from the results reported in~\cite{emervsivc2017unconstrained} for this experiment.      
\begin{table}[!tb]
\caption{Comparative results of the submitted approaches calculated over the entire UERC test dataset (involving $3,540$ subjects). The results are sorted according to rank-1 performance. U17 denotes approaches from UERC 2017.}
\label{tab:scale}
\centering
\footnotesize
\setlength\tabcolsep{5.5pt}

\begin{tabular}{ll rr} 
\hline
Approach & HC/LR$^{\dagger}$ & Rank-1 [\%] & Rank-5 [\%] \\ 
\hline 
\rowcolor{claC} ScNet-5		& Both   & $28.7$	& $39.9	$		\\ 
\rowcolor{claC} CHP-Fus		& Both & $20.5$	& $30.6	$	\\ 
\rowcolor{claC} CH-Fus		& Both & $17.2$	& $27.3	$	\\ 
\rowcolor{claB} MLE-CNN			& LR   & $17.0$	& $29.1	$	\\ 
\rowcolor{claC} LC-Fus	  & Both & $11.1$	& $20.0	$		\\ 
\rowcolor{claB} SEP		& LR & $5.1 $	& $12.6	$		\\ 
\rowcolor{claB} VGGEar		& LR   & $2.4 $	& $6.7 	$		\\ 
\rowcolor{claA} PHOG		& HC   & $2.4 $	& $5.9 	$		\\ 
\rowcolor{claA} IFLBP		& HC   & $0.8 $	& $2.7 	$		\\ 
\rowcolor{claB} SiamCNN		& LR   & $0.5 $	& $1.8 	$		\\ 
\rowcolor{claA} BIOR		& HC   & $0.1 $	& $0.5 	$		\\ 
\hline             	                         	            
\rowcolor{claB} VGG-Base	& LR   & $5.0 $	& $12.0$		\\ 
\rowcolor{claA} LBP-Base	& HC   & $4.5 $	& $10.0$		\\ 
\hline                                                      
\rowcolor{claA} U17 UCCS  & HC  & $23.3$ & $29.0$ 		\\ 
\rowcolor{claC} U17 ITU-II    & Both  & $10.8$ & $20.6$ 		\\ 
\rowcolor{claB} U17 ITU-I   & LR    & $10.1$ & $20.4$ 		\\ 
\rowcolor{claB} U17 IAU     & LR    & $11.2$ & $20.6$ 		\\ 
\rowcolor{claB} U17 IITK    & LR    & $6.7 $ & $11.8$ 		\\ 
\rowcolor{claB} U17 ICL     & LR    & $1.0 $ & $2.8 $ 		\\ 
\hline

\multicolumn{4}{l}{$^{\dagger}$\colorbox{claA}{HC - hand-crafted,} \colorbox{claB}{LR - learned,} \colorbox{claC}{Both (hybrid, LR/HC)}}

\end{tabular}\vspace{-3mm}
\end{table}

\begin{figure}[t!]
    \includegraphics[width=\columnwidth]{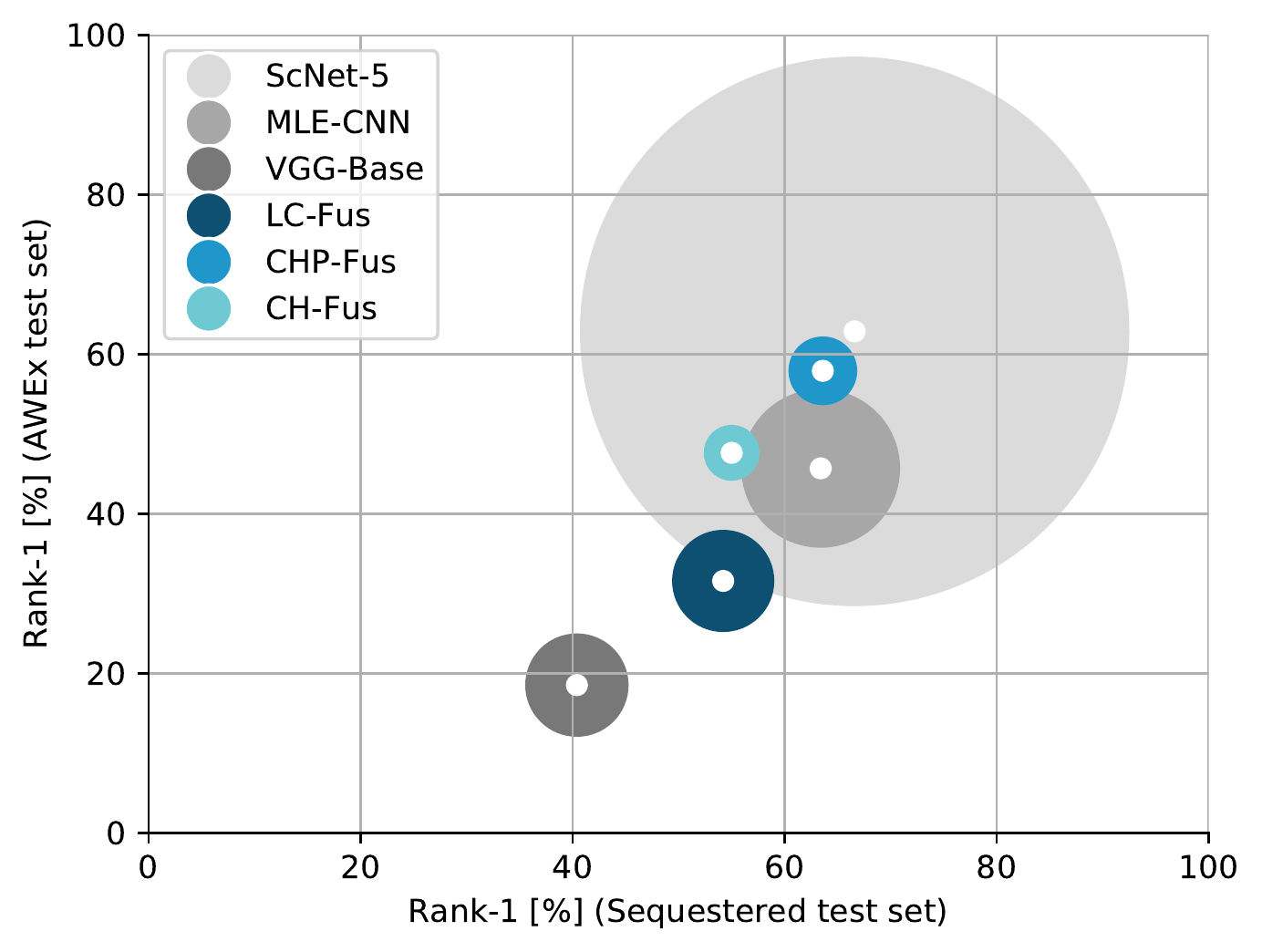} \vspace{-6mm}
    \caption{Assessment of the generalization capabilities of the top $5$ performing techniques from UERC 2019. The $x$-axis shows rank-1 scores for AWEx, the $y$-axis shows the rank-1 results for the sequestered data and circular area represents the model size. Best viewed in color.}\vspace{-3mm}
  \label{fig:seq}
\end{figure}

The results, similarly to the initial test, show the superior performance of techniques relying on learned or on a combination of learned and hand-crafted image descriptors. The relative ranking of the tested techniques is for the most part consistent with the results achieved on the AWEx dataset. The best performing approach is again ScNet-5, however, compared to the first series of experiments, performance has dropped from $62.8\%$ at rank one on AWEx to $28.7\%$ on the complete UERC test data. Similar performance degradations (in the range of $2\times$ to $3\times$ in terms of rank-1 values) are also observed for all other methods. This suggests that, on the one hand, the submitted methods do not scale well with the number of subjects in the test set, but, on the other hand, also shows that the images of the UERC datasets are more challenging due to poorer resolution and lower-quality when compared to AWEx - this is also supported by the results of the sensitivity analysis presented in the next section.    

When comparing this year's submissions to the techniques from UERC 2017, the results show clear improvements. ScNet-5 outperforms the winning approach from 2017 (marked U17 UCCS) in terms of rank-1 score, while CHP-Fus is slightly behind with respect to the rank-1 value, but on pair in terms of the rank-5 result. Four of the submission from this year also outperform the techniques ranked second, third and fourth in 2017. 


\textbf{Experiments on the sequestered dataset.} In the last series of comparative experiments, we evaluate the $5$ top performing algorithms from the previous experiments on a sequestered dataset (with $500$ images and $50$ subjects) that was not available to the participants during development. The idea of this series of experiments is to test the generalization capabilities of the submitted techniques. In Figure~\ref{fig:seq}, where the results are presented, the $x$-axis shows the rank-1 results achieved on the AWEx dataset, the $y$-axis shows the rank-1 scores on the sequestered dataset and the circular surface area around each point represents the size of each model (for exact model size numbers see Table~\ref{tab: approach summary}). The results show that all tested techniques generalize well and achieve slightly better results on the sequestered dataset, which, as discussed earlier, features images of better quality than AWEx due to the manual data collection process. The best performing approach is again ScNet-5, but as can be seen the high performance comes at the expense of a significant model size, which is $2$ orders of magnitude larger than that of the runner up, the CFP-Fus model. 

\subsection{Sensitivity to covariates}

Next, we assess the sensitivity of the submitted models to various factors that are labeled in the AWEx dataset. Specifically, we investigate the impact of occlusion (Figure~\ref{fig:covOcclusion}) and head rotation in different directions (Figures~\ref{fig:covPitch} to~\ref{fig:covYaw}) on the performance of the submitted approaches using the $1800$ test images from the AWEx dataset. 
Furthermore, we also examine the impact of image resolution on the recognition performance in Figure~\ref{fig:covResolution} and use the complete UERC test dataset for this experiment. 
In all graphs, performance is reported in terms of rank-1 recognition scores. 

We see from the presented results that occlusion has a considerable impact on recognition performance. While the rank-1 scores remain relatively unaffected for minor and mild occlusions (i.e. smaller ear accessories or hair covering smaller parts of the ears), severe occlusions (i.e., hair or ear accessory covering around half of the ears) cause considerable performance degradations for the majority of tested techniques. Interestingly, the SEP and VGGEar models are the only ones that seem to ensure stable performance across all degradation levels, but still result in lower rank-1 scores than the top performers with the most severely occluded images. These results suggest that existing recognition techniques are still sensitive to larger occlusions and that novel mechanisms are needed to make ear recognition technology robust toward this specific covariate.
\begin{figure}[tb]
\centering
\hspace{-2mm}
  \includegraphics[width=0.9\columnwidth]{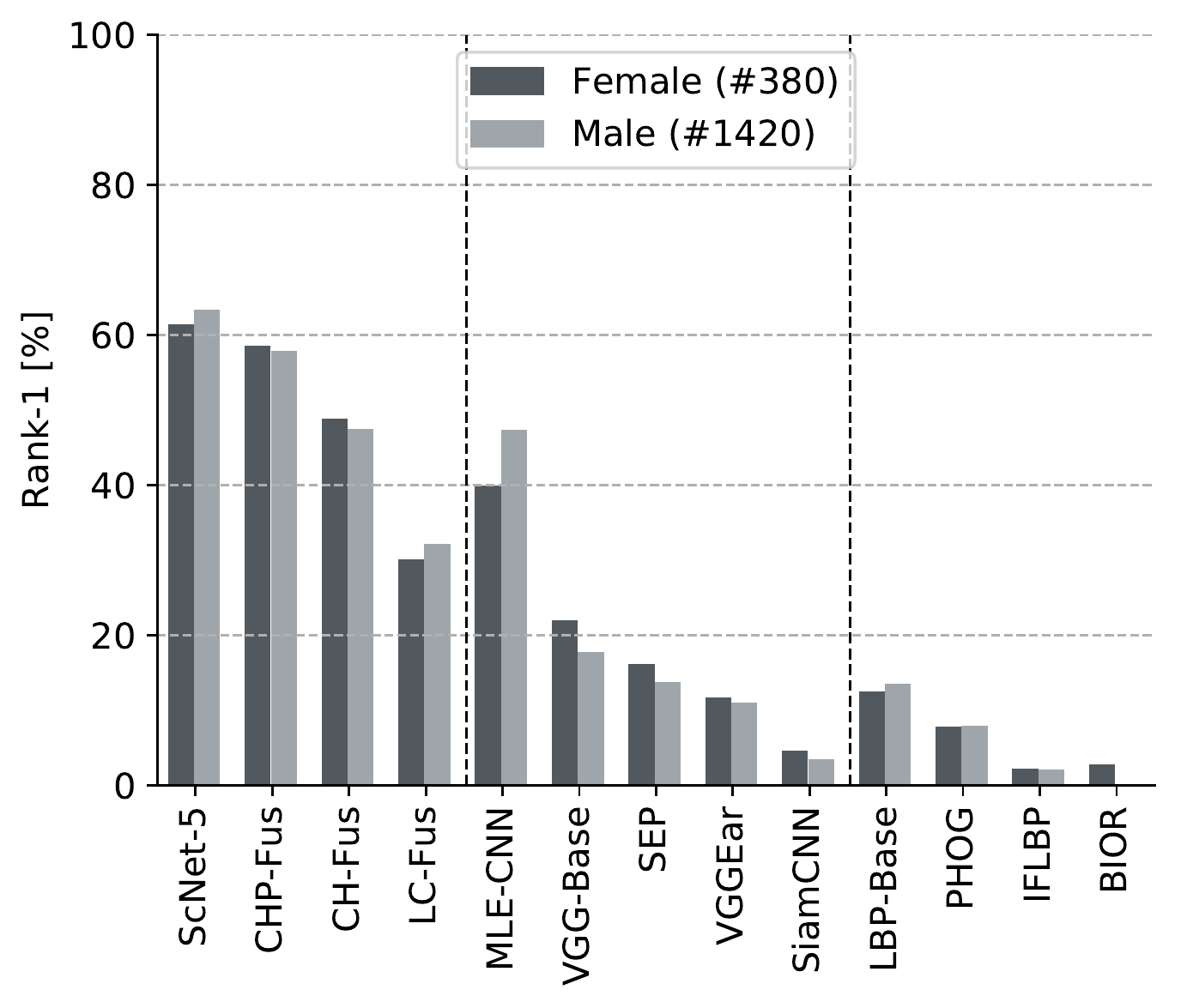}\vspace{-2mm}
\caption{The impact of gender on recognition performance. The bar-graph shows the rank-1 scores for the tested techniques. The number in the brackets indicates the number of images for each gender. All approaches were first grouped by their type (hand-crafted, learned or hybrid) and then sorted according to their rank-1 performance. The vertical delimiters separate groups.}\vspace{-3mm}
\label{fig:covGender}
\end{figure}

Another major source of performance variability on the UERC test data is image resolution. As can be seen in Figure~\ref{fig:covResolution}, all tested methods deteriorate significantly in performance with decreased resolution.
On the smallest images with less than $1$k pixels (e.g. images with a size of $40 \times 25$ pixels), the performance drops by at least a factor of $3\times$ compared to the experiments with the AWEx dataset for all tested approaches.  
Using images with $1$k to $5$k pixels leads to performance degradations of around $2\times$ compared to the AWEx results, while images with a resolution between $5$k pixels and $10$k pixels ensure very similar results to those obtained on AWEx. Results achieved with these images are comparable to the results with images having more than $10$k pixels. This suggests that images of at least $5$k pixels are needed to cater to today's recognition techniques for close to optimal performance, while smaller images inevitably lead to performance degradations.


When looking at the impact of head rotation, we observe very different behaviour for the tested methods. While larger pitch angles lead to significant performance drops for the ScNet-5, CHP-Fus, CH-Fus and LC-Fus and also adversely affect the hand-crafted descriptor-based methods, they have no effect on the MLE-CNN, VGG-Base and SEP models, which even manage to perform slightly better with large pitch angles. Similar observations can be made for variations in roll and yaw angles. The overall top performer, ScNet-5, for example, appears to be unaffected by changes in roll angles, but drops in performance with increased yaw rotations. CH-Fus and LC-Fus, on the other hand, are robust to  yaw rotations, but adversely affected by increased roll angles. MLE-CNN is the most robust among all evaluated approaches in terms of rotation in any direction. While rotation is clearly problematic for many of the tested techniques, it is still less of an issue than occlusion or resolution, as many of the submitted approaches are able to cope with different levels (and directions) in head rotation. 

\subsection{Bias analysis}

In our final series of experiments we test the submitted approaches for performance bias and again use the AWEx test images for the assessment. With this experiment we  try to evaluate if the submitted approaches favor women over men in terms of recognition performance. Since women often wear the same accessories or have longer hair that partially occlude the ears, these factors may contribute towards different recognition rates for male and female subjects. 

As can be seen from  Figure~\ref{fig:covGender}, where the results of this experiment are presented, most approaches ensure similar performance for men and women, except for MLE-CNN, which achieves somewhat higher results for images of male ears. The rest of the tested techniques ensures comparable rank-1 scores and does not favor one gender over the other. 

\section{Conclusion}\label{Sec: conclusion}

The aim of the second Unconstrained Ear Recognition Challenge (UERC 2019) was to evaluate the current state of ear recognition technology and assess the progress made in the field since the first competition in 2017. While ma\-ny open issues still remain, rank-1 recognition rates on the most challenging (large-scale) experiment improved by more than $5\%$ from 2017, while rank-5 recognition rates improved by more than $10\%$ considering the top performers of the two challenges.

UERC 2019 was used as a platform to evaluate different aspects of ear recognition techniques, such as sensitivity to image resolution, occlusion, head rotation and performance bias with respect to gender. While the submitted approaches have in general become more competitive compared to the 2017 submissions, they are still sensitive to ear occlusion and insufficient image resolution. Head rotations and gender bias, on the other hand, appear to be less problematic.

For future challenges we plan to include additional datasets and especially training data more suitable for learning CNNs, which seem to be the favored and best performing models in the field of ear recognition.

\section*{Acknowledgments}

The research presented in this paper was supported in parts by the ARRS (Slovenian research agency) Research Programme P2-0250 (B) Metrology and Biometric Systems and the ARRS  Research Programme P2-0214 (A) Computer Vision. The authors would also like to thank the Nvidia corporation for donating many of the
GPUs used in this work.

{\small
\bibliographystyle{ieee}
\bibliography{uerc2019}
}

\clearpage

\appendix{}

\section{Appendix -- Summary of Approaches}

 In this part of the appendix we present a short summary of all approaches participating in UERC 2019. We first describe the two baselines provided by the organizers of UERC and then discuss the participating approaches.

\subsection{LBP-baseline (LBP-Base)}

{\noindent {\small  {\bf Contributors}: 
Ž. Emeršič$^1$,  L. Yuan$^2$, H.~K. Ekenel$^3$, P. Peer$^1$, V.~Štruc$^1$ (UERC 2019 organizers)}}\\
{\noindent \small \it  $^1$University of Ljubljana (UL), Slovenia\\ $^2$University of Science and Technology Beijing (USTB), China \\$^3$Istanbul Technical University (ITU), Department of Computer Engineering, Turkey\\}

{\noindent 
LBP-Base is the first baseline included in the UERC 2019 starter kit made available to the UERC participants. The baseline represents a descriptor-based approach that relies on histograms of local binary patterns (LBPs)~\cite{pietikainen2011local}. With this baseline, $16\times 16$ patches are sampled from the ear images using a sliding window approach and a step size of $4$ pixels. Each patch is then encoded with uniform LBPs computed with a radius of $R=2$ and a local neighborhood size of $P=8$. $59$-dimensional histograms are calculated for every patch and the computed histograms are finally concatenated to form the final ear descriptor. The cosine similarity is used to measure similarity between two ear (LBP) descriptors.} 

\subsection{VGG-baseline (VGG-Base)}

{\noindent \small  {\bf Contributors}:  Ž. Emeršič$^1$,  L. Yuan$^2$, H.~K. Ekenel$^3$, P. Peer$^1$, V.~Štruc$^1$ (UERC 2019 organizers)\\}
{\noindent \small \it $^1$University of Ljubljana (UL), Slovenia\\ $^2$University of Science and Technology Beijing (USTB), China \\$^3$Istanbul Technical University (ITU), Department of Computer Engineering, Turkey\\}

{\noindent
VGG-Base is the second baseline included in the UERC 2019  starter kit. This baseline implements the popular $16$-layer VGG convolutional neural network (CNN) from~\cite{simonyan2014very} for ear recognition. The model consists of multiple convolutional and max-pooling layers and  uses small filters (of size $3\time 3$ pixels) to reduce the number of parameters that need to be learned during training. The convolutional layers are followed by two fully-connected layer and the output of the second fully-connected layer is used as a $4,096$-dimensional ear descriptor. The VGG baseline is trained from scratch using the UERC training data and aggressive data augmentation. A detailed description of the training procedure is presented in~\cite{ziga@bwild}. Two ear descriptors are again matched using the cosine similarity.}

\subsection{Pyramid Histogram of Oriented Gradients (PHOG)}

{\noindent \small {\bf Contributors}: A.~Kumar S.~V.$^1$,  B. S. Harish$^2$\\ }
{\noindent \small \it $^1$Nitte Mahalinga Adyanthaya Memorial Institute of Technology (NMAM IT), India\\$^2$Jagadguru Sri Shivarathreeshwara Science \& Technology University (JSS STU), India\\}

{\noindent 
This approach uses the PHOG (Pyramid of Histograms of Oriented Gradients) descriptor from~\cite{bosch2007representing} to extract features from grey-scale ear images. PHOG is a spatial shape descriptor, which represents ear images by their local shape, but different from competing image descriptors preserves the spatial information of the encoded ear shape. To compute the PHOG descriptor, grey-scale ear images are encoded at multiple (pyramid) levels using the standard HOG (Histogram of Oriented Gradients) descriptor. For each encoding a grid is defined over the image containing a predefined number of cells. At the first level, the entire images is considered a single cell, at the second level the image is partitioned into four cells and at the third level the image is partitioned into a total of 16 cells. 
Finally, the PHOG descriptor for an image is computed by concatenating the HOG descriptors from all grid-pyramid levels. The cosine similarity is used for scoring.}

\subsection{Intuitionistic Fuzzy Local Binary Pattern (IFLBP)}

{\noindent \small  {\bf Contributors}:  A.~Kumar S.~V.$^1$,  B. S. Harish$^2$\\ }
{\noindent \small \it $^1$Nitte Mahalinga Adyanthaya Memorial Institute of Technology (NMAM IT), India\\$^2$Jagadguru Sri Shivarathreeshwara Science \& Technology University (JSS STU), India\\}

{\noindent 
IFLBP uses Intuitionistic Fuzzy Local Binary Pattern (IFLBP)~\cite{ansari2016feature} to extract features from grey scale ear images. IFLBP represents a variant of the conventional Fuzzy LBP operator, which overcomes the limitations the original LBP, such as hard thresholding and sensitivity towards the gray level uncertainty in the input image. However, Fuzzy LBP fails to address another uncertainty, which arises when defining the membership function for the descriptor. IFLBP overcomes this limitation by considering both membership and non-membership values. To compute IFLBP, the grey level ear image is ``fuzzified'' using a triangular membership function. Next, Fuzzy LBP values are computed for each pixel. Finally, the membership value associated with each pixel is updated by adding a hesitation degree. Image descriptors are compared using the cosine similarity.}

\subsection{SiamEarPers (SEP)}
{\noindent {\small {\bf Contributors}: W. Gutfeter$^1$,  J.~N. Khiarak$^2$, A. Pacut$^2$\\}}
{\noindent {\small \it  $^1$Research an Academic Computer Network (NASK), Poland\\ $^2$Warsaw University of Technology (WUT), Poland \\}}
 
 {\noindent 
The SiamEarPers (SEP) model is based on a Siamese neural network, where the residual network architecture ResNet-50~\cite{he2016deep} was used as the backbone model. SEP uses a hybrid loss function that consists of a classification-learning component (softmax loss function) and a distance-learning component (Siamese loss function). The novelty of this approach lays in the concept of training the first function with respect to object-dependent features (specific to ears) and the second part with respect to features specific to a person (the user). Images are resized to an input resolution of $112\times 80$. No preprocessing is done before feeding the images to the model. To compare images, the cosine distance  between $2048$-dimensional ear descriptors extracted by the SEP model is used.}
 

\subsection{CNN+HOG (CH-Fus)}
{\noindent \small {\bf Contributors}: E. Hansley$^1$,  M. Pamplona Segundo$^2$, S. Sarkar$^1$\\}
{\noindent \small \it  $^1$University of South Florida (USF), USA\\ $^2$Federal University of Bahia (UFBA), Brasil \\}

 {\noindent 
The CH-Fus approach automatically normalizes, describes and matches grayscale images of cropped ears~\cite{hansleyEmployingFusionLearned2018}. For normalization, a CNN-based side detector is used to flip the image horizontally if it is left-oriented and a CNN-based landmark detector is exploited to locate a set of 55 landmarks, which are employed to translate, rotate and scale an input image to a standard configuration. To this end, a two-stage landmark detector was designed that produces accurate results even in the presence of extreme variations that were not seen during training. For description, features learned by a CNN are combined with handcrafted features extracted by HOG, which produce 512- and 8712-dimensional vectors and are matched by cosine and chi-squared distances, respectively. Before fusion, score normalization is carried out considering an identification scenario, in which the only scores available at a single time are the ones between the probe and all gallery images.  Min-max normalization is performed and the obtained scores are fused with the sum rule. The landmark detector employed is trained using the In-the-wild Ear Database~\cite{Ref2} and the side detection and description networks is trained using the UERC training set.}
 


\subsection{CNN+HOG+POEM (CHP-Fus)}

{\noindent \small {\bf Contributors}:  M. Pamplona Segundo$^1$, S. Sarkar$^2$\\}
{\noindent \small \it  $^1$Federal University of Bahia (UFBA), Brasil\\ $^2$University of South Florida (USF), USA \\}

 {\noindent 
 The CHP-Fus approach automatically normalizes, describes and matches grayscale images of cropped ears. Normalization is carried out by Hansley et al.'s approach~\cite{hansleyEmployingFusionLearned2018}. For description, learned features extracted by Hansley et al.'s CNN are combined with handcrafted features extracted by HOG and POEM, which produce 512-, 8712- and 11328-dimensional vectors, respectively. A second CNN is trained, hereon referred to as CNN+, using Hansley et al.'s architecture and additional training data: all images from Indian Institute of Technology Delhi Ear~\cite{Kumar2012}, West Pomeranian University of Technology Ear~\cite{Frejlichowski2010} and In-the-wild Ear~\cite{Ref2} databases together with UERC training images. Multilayer perceptrons (MLP) are also trained for HOG and POEM features to transform them into 512-dimensional vectors using this extended training set, hereon referred to as HOG+ and POEM+. HOG and POEM features are matched by chi-squared distance while CNN and MLP features are matched by cosine distance. The scores of these six features (CNN, HOG, POEM, CNN+, HOG+ and POEM+) are combined using the sum rule after performing a min-max normalization. The main difference to Hansley et al.'s fusing scheme (i.e., CH-Fus) is that CHP-Fus first transforms each feature to its enhanced version  and then combines all features together (e.g. CNN+CNN+).}
 


\subsection{Ensemble of four CNNs (MLE-CNN)}

{\noindent \small {\bf Contributors}:  H. Park$^1$, G. Pyo Nam$^1$, I.~J. Kim$^1$\\}
{\noindent \small \it  $^1$Korea Institute of Science and Technology (KIST), South Korea\\}

 
 {\noindent 
 The MLE-CNN approach ensembles 4 CNN models, i.e., two VGG-16 and two ResNet-50 models, all pretrained on the ImageNet database. The four models are grouped into two pairs, where each pair consists of a VGG and a ResNet model. Since all of the model use $224\times224$ color images as input, all images are resized to $224 \times 224$ pixels, while maintaining the original aspect ratio with zero padding. Because of insufficient training data in UERC, data augmentation is performed and 46,378 images are generated by cropping, flipping, etc. The dataset is divided into a training set and a validation set in an 8:2 ratio. During the learning stage, the first network pair is trained directly with the augmented UERC dataset. For the second network pair, an additional database, i.e.,  K-Ear dataset, is used for domain adaptation before using UERC. For the ResNet models, the original ReLU activations are replaced by better performing Leaky ReLU activation functions. The output of the second fully connected layer of the VGG-16 models and averaged positive values of the final convolutional layer of the ResNet models are selected as features for the input ear images. To consider both sides of the ear (left and right) a feature vector from a flipped version of the input image is added to the original one. The cosine similarity is calculated to compare feature vectors of two ear images. Scores from the four CNNs are combined based on a weighted SUM in order to enhance the recognition performance. }

\subsection{Twin Siamese Network (SiamCNN)}

{\noindent \small {\bf Contributor}:  S.~G. Sangodkar$^1$\\}
{\noindent \small \it  $^1$Indian Institute of Technology Bombay (IITB), India\\}

 {\noindent
 The SiamCNN approach is based on a Siamese architecture involving a twin Resnet-18 CNN network. Initially, the USTB ear image dataset is used to perform transfer learning on a Resnet-18 CNN model pretrained on Imagenet data. The (transfer) learned model is then used in the Siamese architecture to train on randomly sampled pairs (positive-positive and positive-negative) of images from the UERC training dataset. Care is taken to ensure that an equal number of positive-positive and positive-negative pairs are generated. The contrastive loss function based on the Euclidean distance measure and the margin of $5$ is used. During training, the network takes two color  ear images resized to $224\times224$ pixels as input and passes them through the twin network. The Euclidean distance between the two feature vectors (dimension: $2304$) corresponding to the pair of input images is evaluated and the corresponding loss is calculated using the contrastive loss function, which is then used to simultaneously update the weights of the twin CNN network. }

\subsection{ScoreNet-5 (ScNet-5)}

{\noindent \small {\bf Contributors}:   U. Kacar$^1$, M. Kirci$^1$\\}
{\noindent \small \it  $^1$Istanbul Technical University (ITU), Department of Electronics and Communication Engineering, Turkey\\}


{\noindent 
ScoreNet-5~\cite{kacarScoreNetDeepCascade2018} represents the first automated fusion-learning (AutoFL) approach for ear recognition.
Building the ScoreNet model includes three basic steps. In the first step, a diverse and modular pool of modalities is created. The term modality in this context refers to different operations along the standard recognition pipeline, e.g., image resizing (with different target sizes), preprocessing (with different technqiues), image representation (using hand-crafted or learned features), dimensionality reduction (with multiple options), and distance measurement (with various distance measures). In the second step, random modalities are selected for each processing step resulting in a randomly generated ear recognition pipeline. The pipeline is then spplied on a training and validation set to generate a similarity score matrix. Once multiple such pipelines are generated, a Deep Casaceded Score Level Fusion (DCSLF) algorithm selects the best groups of modalities (i.e., pipelines) and calculates the necessary fusion weights. In the third step, all selected modality groups (pipelines) and calculated fusion weights in the cascaded network structure are fixed for the test dataset~\cite{kacarScoreNetDeepCascade2018}.
For UERC 2019, the landmark-based orientation and scale-normalization procedures  from~\cite{hansleyEmployingFusionLearned2018} are implemented and applied on the   ear images of the experimental database before feeding them to ScoreNet. The ScoreNet model is trained with DCSLF using only the training part of the AWEx dataset and contains a total of $41$ modality groups (pipelines). In terms of image representations, ScoreNet combines HOG, Gabor, uLBP, LPQ and BSIF descriptors as well as AlexNet, VGG-16, VGG-19, GoogLeNet, ResNet-101, ResNet-18, SqueezeNet, InceptionV3, IncetionResNetV2 and DenseNet models.}




\subsection{VGG Ear (VGGEar)}

{\noindent \small {\bf Contributors}:   L.~Yuan$^1$, J.~Yuan$^1$, H.~Zhao$^1$, F.~Lu$^1$, J.~Mao$^1$,  X.~Zhang$^1$\\}
{\noindent \small \it  $^1$University of Science and Technology Beijing (USTB), China\\}

{\noindent 
VGGEar relies on multiple VGG-like models for ear image representation. In this approach, the standard VGG model is first modified and then fine-tuned  on images from the USTB-Helloear database~\cite{zhangEarVerificationUncontrolled2018}. First, the last pooling layers are replaced by spatial pyramid pooling layers to accomodate arbitrary data sizes and obtain multi-level features. In the training phase, the VGG model is trained both under the supervision of the softmax and center losses to obtain more compact and discriminative features to identify unseen ears. Finally, three VGG models with different scales of ear images are assembled to generate multi-scale ear representations for ear description. The cosine similarity is used to measure the similarity between two ear descriptors.}


\subsection{CNN\&LBP (LC-Fus)}
{\noindent \small {\bf Contributors}:   D.~Yaman$^1$, F.~I.~Eyiokur$^1$, K.~B.~\"{O}zler$^1$, H.~K.~Ekenel$^1$\\}
{\noindent \small \it  $^1$Istanbul Technical University (ITU), Department of Computer Engineering, Turkey\\}

{\noindent 
The LC-Fus method benefits from both hand-crafted and CNN based features. For the CNN part, the VGG-16~\cite{simonyan2014very} model is selected and a combination of center loss~\cite{wen2016discriminative} and softmax loss is used to learn the model parameters. 
While the softmax loss tries to produce separable features for each class, center loss is responsible of producing discriminative features by using the total distance between features and the corresponding class center. In order to improve the discriptivness of the feature representation, a domain adaptation strategy is also utilized. For this, instead of initializing network parameters with the pretrained models trained on the ImageNet dataset~\cite{dengImagenetLargescaleHierarchical2009}, a CNN model fine-tuned on the ear domain is used. To provide domain adaptation, the VGG-16 model is fine-tuned on the extended version of the Multi-PIE ear dataset~\cite{eyiokurDomainAdaptationEar2017}. Afterwards,  this model is further fine-tuned on the UERC 2019 training data. Data augmentation is performed on the UERC 2019 training data to avoid over-fitting. For the evaluation part,  features from the FC6 layer of the final VGG-16 model are extracted and the similarity matrix is calculated with the cosine metric. For the hand-crafted fetures, the LBP baseline~\cite{pietikainen2011local} is used in the LC-Fus approach. The LBP features are normalized and the similarity matrix is again computed with the cosine metric. Finally, both similarity matrices are combined to evaluate performance of the LC-Fus model.}






\subsection{Bior4.4-Energy (BIOR)}

{\noindent \small {\bf Contributors}:   D.~Paul~Chowdhury$^1$, S.~Bakshi$^1$, P.~K.~Sa$^1$, B.~Majhi$^1$\\}
{\noindent \small \it  $^1$National Institute of Technology Rourkela (NITR), India\\}





{\noindent 
The BIOR approach~\cite{chowdhury2018applicability} uses energy features computed based on the fourth-level biorthogonal-tunable-wavelet (Bior 4.4) image decomposition to represent ear images for recognition. Since part of the ear image may be occluded by hair or spectacles, local energy features are considered by subdividing images into six blocks and computing the wavelet decomposition for each block. Then, the energy of each decomposed block is calculated and used as a feature for the corresponding ear image. The size of the final feature vector is $72$. The cosine distance is used to measure the similarity between two feature vectors. The extracted feature vector is found to work on par with level-2 or level-3 energy coefficients as well as with the entire set of wavelet coefficients. }




\section{Appendix -- Qualitative Evaluation}

\begin{figure*}[!tb]
 \centering
 \includegraphics[width=0.9\textwidth]{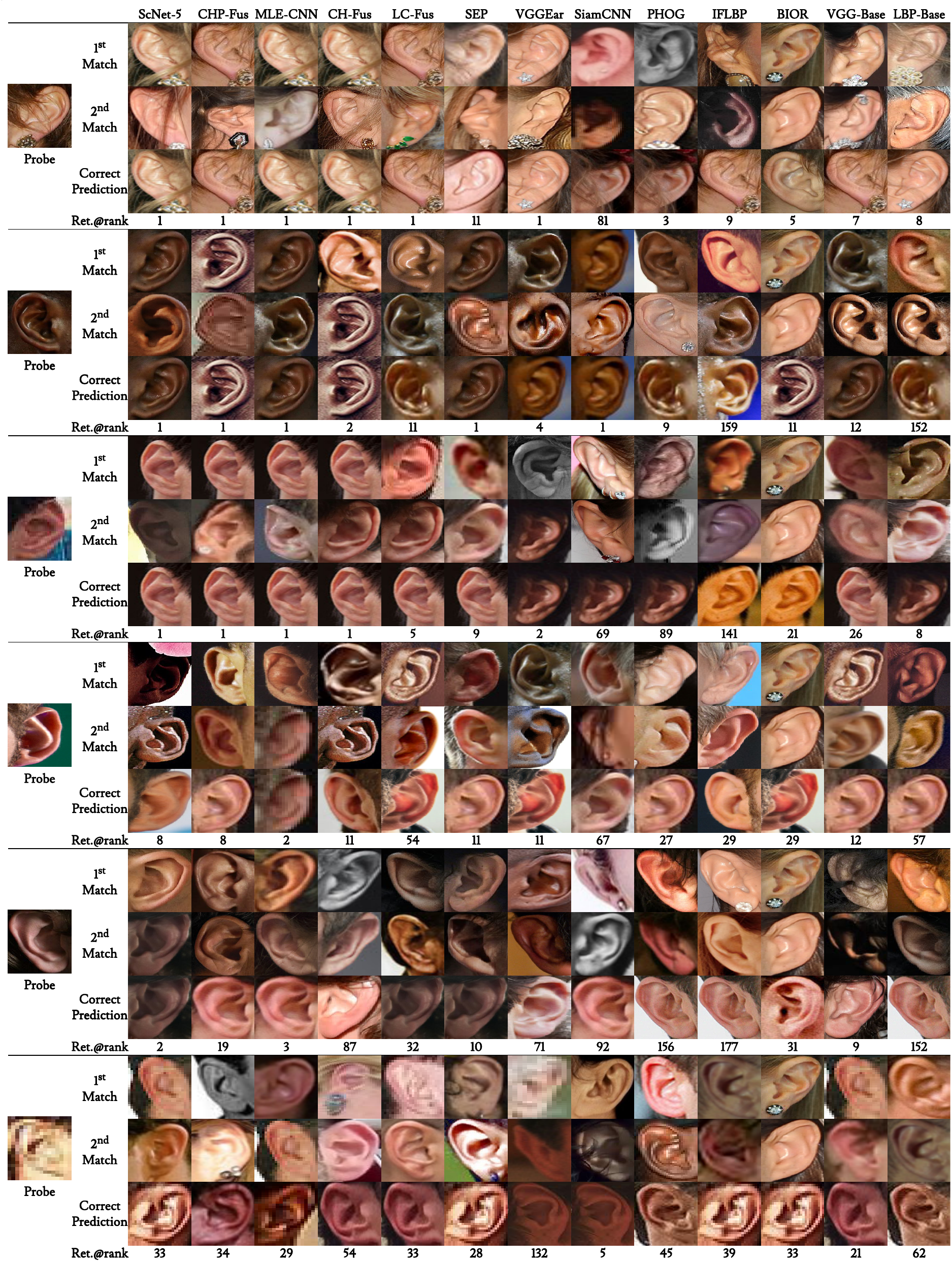}
\caption{Qualitative analysis with selected probe images. The figure shows selected probe images (on the left) and the first and second match generated by the submitted approaches. The first retrieved image with the correct identity is also shown together the corresponding rank, at which it was retrieved. 
}
\label{fig:qualitative}
\end{figure*}

In this part of the Appendix we present qualitative results and show what type of images the submitted approaches return as the first (rank-1) and as the second match (rank-2) for a given probe images. We also show the first correct prediction (note that there are multiple images of the correct subject in the gallery) and provide the rank, at which it was retrieved. The first correct prediction is considered to be the image that is closest in the ranking and has the same identity as the probe image. This qualitative analysis is shown in Figure~\ref{fig:qualitative} for 6 randomly selected probe images - shown on the left.

The qualitative results reflect the quantitative analysis well. With the top performing approaches, the images retrieved at rank 1 and 2 exhibit a high visual similarity to the probes, as expected. Thus, even when predictions fail, the closest matches visually resemble the probe image. As already indicated by the sensitivity analysis in the main paper, image resolution is problematic. This is illustrated with the 6th probe image at the bottom of the figure, where all submitted approaches fail to retrieve the correct identity. Nevertheless, in terms of appearance, many of the tested technqiues again retrieve visually similar results. 

\end{document}